\documentclass[12pt,a4paper]{article}

\usepackage[utf8]{inputenc}
\usepackage[T1]{fontenc}
\usepackage{amsmath,amssymb,amsthm}
\usepackage{mathtools}
\usepackage{booktabs}
\usepackage{enumitem}
\usepackage{hyperref}
\usepackage[margin=1in]{geometry}
\usepackage{graphicx}
\usepackage{natbib}
\usepackage{url}
\usepackage{float}
\usepackage{caption}

\newtheorem{theorem}{Theorem}[section]

\newtheorem{observation}[theorem]{Observation}
\theoremstyle{definition}
\newtheorem{definition}[theorem]{Definition}

\newcommand{\SCS}{\textrm{SCS}}

\newcommand{\CSTS}{\textrm{C-STS}}
\newcommand{\vect}[1]{\mathbf{#1}}
\newcommand{\norm}[1]{\left\|#1\right\|}
\newcommand{\card}[1]{\left|#1\right|}

\title{Semantic Conditional Similarity (SCS):\\A Formal Mathematical Framework}
\author{Mehmet Murat Albayrakoglu\\
\small Işık University, Department of Management Information Systems\\
\small Istanbul, Turkey\\
\small \texttt{murat.albayrakoglu@isikun.edu.tr}}
\date{}

\begin{document}

\maketitle

\begin{abstract}
Semantic Textual Similarity (STS) operates on document pairs, returning a single scalar for each comparison. In multi-document settings where several source texts may have contributed to a single target, pairwise STS cannot disentangle the marginal semantic contribution of each source from the vocabulary they share. This paper introduces Semantic Conditional Similarity (SCS), a conditional form of STS in which the similarity between two documents is computed after eliminating the semantic elements shared with a third document, via set-theoretic operations. SCS makes four contributions. First, it introduces \emph{directionality} to semantic similarity, partitioning pairwise relations into lateral (inter-source) and vertical (source-to-target) classes. Second, the conditioning operation induces \emph{role-dependent asymmetry}: extending the comparison domain from unordered document pairs to ordered, conditioned triples produces scores that depend on which source serves as reference and which as conditioning document. The asymmetry is a consequence of the set-theoretic construction, not a relaxation of the similarity function itself. Third, SCS introduces \emph{inter-document conditioning}, which subtracts shared content from a third document, as a complement to the intra-document attribute-based conditioning of recent Conditional STS (C-STS) work. Fourth, the set-theoretic decomposition achieves \emph{threshold-free partitioning} of shared and residual content, requiring no arbitrary similarity cutoff. Although developed for the many-to-one ($n \to 1$) configuration, the framework extends to the one-to-many ($1 \to n$) case through a vertical role reversal of the conditioning structure, and to the many-to-many ($m \to n$) case through the independent composition of source-side and target-side conditioning. Together with the established one-to-one ($1 \to 1$) case, these extensions span the entire document cardinality spectrum. SCS scores should be interpreted ordinally, as rankings of marginal contributions rather than as absolute measurements. The broader implication is that the metric space axioms commonly assumed in similarity formalizations impose substantive constraints on expressiveness in role-differentiated multi-document settings, and that extending the comparison domain to conditioned triples can recover information that pairwise-symmetric comparisons suppress.
\end{abstract}

%------------------------------------------------------------------
\section{Introduction}
\label{sec:intro}

\subsection{The problem}

Semantic Textual Similarity (STS) is a cornerstone of Natural Language Processing (NLP), information retrieval, and computational linguistics; it measures the degree of overlap between the meanings of two texts \citep{agirre2012semeval}. In its standard formulation, STS operates on document pairs: given two documents, it returns a single scalar that indicates how semantically similar they are. This pairwise design is adequate when the analytical question is itself pairwise: \emph{how similar is document~$A$ to document~$B$?}

Many real-world analytical settings, however, involve a richer structure. As in studies of intellectual influence, legislative drafting, policy diffusion, or intertextual analysis, when a target text is situated within a context of multiple source texts that may have contributed to it, the relevant question is no longer merely \emph{how similar} each source is to the target. Instead, one should ask \emph{how much each source contributed uniquely}, net of what it shares with the other sources. Standard pairwise STS cannot answer this question. If two source documents share vocabulary with the target and with each other, their pairwise STS scores with the target will each be inflated by the shared inter-source context. There is no mechanism within the pairwise framework to disentangle the marginal contribution of each source from the confounding overlap among them.

\subsection{Empirical motivation}

The need for such a mechanism emerged from an earlier study on the semantic relationship between normative theories of ethics and institutional texts. In that study, \citet{albayrakoglu2022influence} applied classical semantic similarity analysis to compare theories of ethics against organizational codes of conduct/ethics, using unconditioned pairwise STS to assess the alignment between each theory and each code. While this approach revealed broad patterns of semantic association, it could not distinguish whether the observed similarity between a given theory and a given code reflected a genuine, unique alignment or was an artifact of vocabulary shared across multiple theories, i.e., vocabulary that would appear in any code regardless of its specific theoretical orientation.

This limitation became more acute in a subsequent work by \citet{albayrakoglu2026semantic}, in which the authors applied transformer-based STS to analyze the semantic alignment between normative theories of ethics and the European Union Artificial Intelligence (EU AI) Act. This is an explicitly multi-source setting: multiple theories (deontological ethics, virtue ethics, and consequentialism) are examined as potential source documents of a single legislative text. When several theories share foundational ethical vocabulary using terms like ``harm,'' ``fairness,'' ``transparency,'' and ``accountability,'' unconditioned pairwise STS assigns elevated similarity scores to all of them without revealing which theory uniquely contributed to the legislative text. The shared ethical lexicon inflates every score, obscuring the differential alignment that is the actual object of inquiry.

\subsection{The response: Semantic Conditional Similarity}

This paper introduces \emph{Semantic Conditional Similarity (SCS)}, a conditional form of STS in which the similarity between two documents is computed after eliminating semantic elements shared with a third document in a given context. SCS addresses the confounding problem described in empirical motivation by conditioning the similarity computation on the inter-source overlap. For each pair of source documents, their shared vocabulary is subtracted from both the source and target representations before similarity is computed, isolating the marginal semantic contribution of each source to the target.

The framework makes the following contributions:

\paragraph{Directionality.} SCS introduces a structural distinction between \emph{lateral} similarity (among source documents) and \emph{vertical} similarity (between a source and the target), partitioning the similarity space of a multi-document configuration into two functionally distinct classes. This is a departure from classical STS, in which all pairwise similarities belong to a single undifferentiated set (Section~\ref{sec:directionality}).

\paragraph{Asymmetry.} By extending the comparison domain from unordered document pairs to ordered, conditioned triples, SCS produces scores that depend on which source serves as reference and which as conditioning document: the marginal contribution of source~$i$ to the target, net of what $i$ shares with source~$j$, is not in general equal to the marginal contribution of $j$ net of what $j$ shares with $i$. The underlying similarity function (cosine) remains symmetric; the asymmetry arises because swapping roles changes the inputs to the function, not the function itself, and hence, it is role-induced  (Section~\ref{sec:asymmetry}).

\paragraph{Inter-document conditioning.} Unlike recent work on Conditional STS (C-STS)  \citep{deshpande2023csts,zhang2026case,song2025polirl}, which conditions on named semantic attributes within the compared texts (intra-document conditioning), SCS conditions on the shared content of a third document outside the pair under comparison (inter-document conditioning). The two forms of conditioning address different problems: intra-document conditioning disambiguates which aspect of similarity to evaluate; inter-document conditioning decomposes multi-document similarity into marginal contributions.

\paragraph{Threshold-free partitioning.} SCS operates via set-theoretic operations---mainly intersection and complement---that yield a lexical decomposition of shared and residual content without requiring an arbitrary similarity threshold. The partition is a set-theoretic fact, not a judgment call.

More broadly, SCS has implications for the foundations of similarity theory: the metric space axioms commonly assumed in similarity formalizations, including the symmetry axiom which is preserved throughout the unified theory of \citet{rozinek2021duality}, encode substantive assumptions about the relationships being modeled. These assumptions become visible once the comparison domain is enlarged beyond unordered pairs (Section~\ref{sec:related}).

Although developed for the many-to-one ($n \to 1$) configuration, the conditioning mechanism extends to the one-to-many ($1 \to n$) case through a vertical role reversal, and to the many-to-many ($m \to n$) case through the independent composition of source-side and target-side conditioning. Together with the established one-to-one ($1 \to 1$) case, these extensions cover the entire document cardinality spectrum (Section~\ref{sec:extensions}).

\subsection{Paper organization}

The remainder of this paper is organized as follows. Section~\ref{sec:foundations} establishes the mathematical foundations. Section~\ref{sec:decomposition} develops the multi-document semantic decomposition. Section~\ref{sec:departures} identifies the two principal departures from classical STS. Section~\ref{sec:bridge} presents the set-theoretic bridge. Section~\ref{sec:scs} introduces the SCS framework, including the algorithmic workflow. Section~\ref{sec:properties} discusses properties and constraints. Section~\ref{sec:extensions} extends the framework across the full document cardinality spectrum. Section~\ref{sec:demonstration} provides a synthetic numerical demonstration. Section~\ref{sec:related} positions SCS with respect to related work. Section~\ref{sec:discussion} discusses broader implications. Section~\ref{sec:conclusion} concludes the paper.

%------------------------------------------------------------------
\section{Foundational Setting}
\label{sec:foundations}

\subsection{Semantic space as a normalized metric space}

Let $X$ be a non-empty set of textual documents. A \emph{normalized metric} on $X$ is a function $\delta : X \times X \to [0,1]$ satisfying, for all $x, y, z \in X$:

\begin{enumerate}
  \item \textbf{Identity:} $\delta(x, y) = 0 \iff x = y$
  \item \textbf{Symmetry:} $\delta(x, y) = \delta(y, x)$
  \item \textbf{Triangle inequality:} $\delta(x, y) \leq \delta(x, z) + \delta(z, y)$
\end{enumerate}

The pair $(X, \delta)$ is a \emph{normalized metric space}. Semantic similarity is defined as the complement of distance $\delta$:
\begin{equation}
  \sigma(x, y) = 1 - \delta(x, y)
\end{equation}
where $\sigma : X \times X \to [0,1]$ is the \emph{semantic similarity function}.

\subsection{Document representation as unit-area regions}

\begin{figure}[t]
	\centering
	\includegraphics[width=0.75\textwidth]{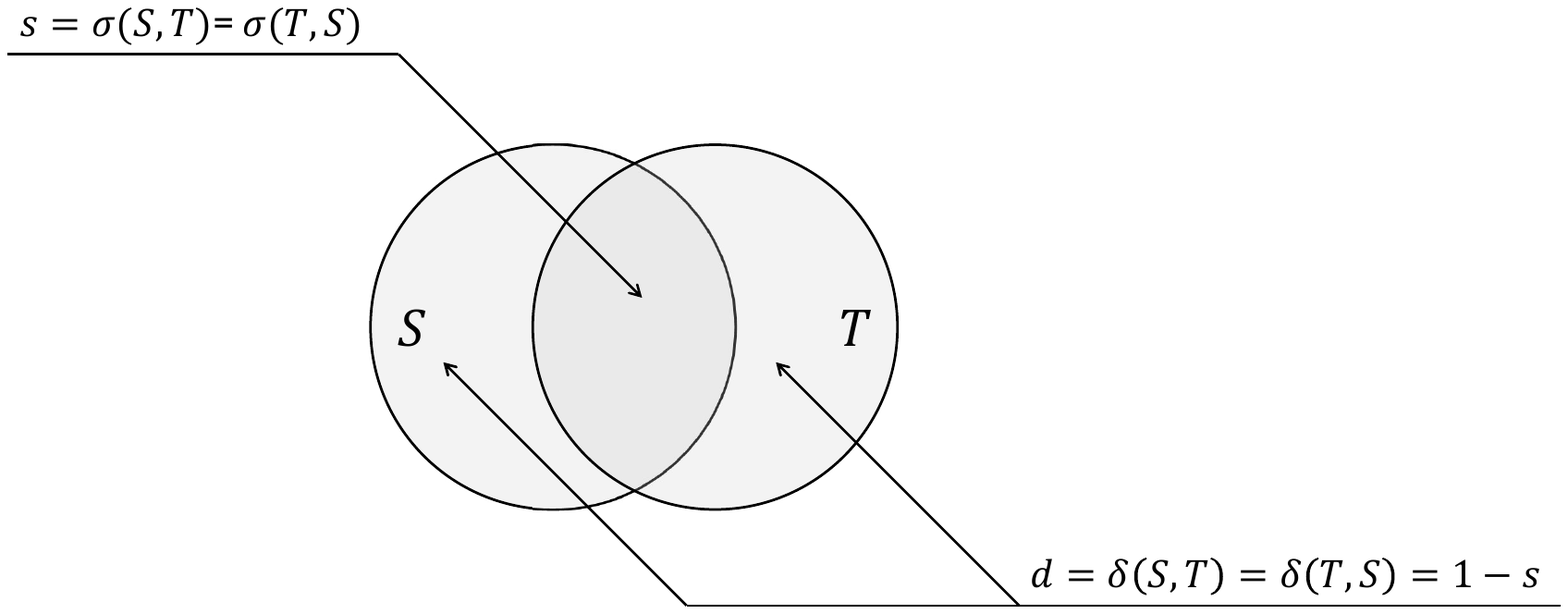}
	\caption{Pairwise similarity as two overlapping normalized documents, $S$ and $T$.  The overlap
		region corresponds to the similarity $s = \sigma(S, T)$; the
		non-overlapping portions correspond to $d = 1 - s$.}
	\label{fig:pairwise}
\end{figure}

Each document is modeled as a \emph{unit-area region} in a conceptual space encapsulating the document's structure and content \citep{gardenfors2000conceptual}. This normalization ensures that comparisons reflect qualitative semantic overlap rather than superficial differences in length, as shown in  Figure~\ref{fig:pairwise}.

\begin{definition}[Unit-area regions]\label{def:unit-area}
	Each document~$d$ is represented as a bounded planar region of unit area. For any two document regions, the area of their overlap equals the similarity~$\sigma$ and the area of their non-overlapping portions equals the dissimilarity~$\delta = 1 - \sigma$. The regions need not be literal geometric circles; any bounded shapes of equal area suffice.
\end{definition}

For one source document $S$ and one target $T$:
\begin{align}
	s &= \sigma(S, T) = \sigma(T, S) & \text{(similarity)} \\
	d &= \delta(S, T) = \delta(T, S) = 1 - s & \text{(dissimilarity)}
\end{align}

Under the unit-area assumption, the normalized symmetric difference between two regions equals $1 - s$ by inclusion-exclusion, so the geometric and metric formalizations are interchangeable, linking the geometric illustration to the framework of \citet{rozinek2021duality}. Throughout this paper, Venn diagrams serve as the visual representation of documents: each circle corresponds to a document's unit-area region, and each distinct zone corresponds to a portion defined by a unique pattern of overlap and non-overlap across the documents. From Section~\ref{sec:bridge} onward, the abstract regions are grounded in concrete lemma sets.

%------------------------------------------------------------------
\section{Multi-Document Semantic Decomposition}
\label{sec:decomposition}

\subsection{Document cardinalities}

Let $\{S_1, S_2, \ldots, S_k\}$ be a set of \emph{source} documents and $T$ be a single \emph{target} document. The framework addresses the $n \to 1$ (many-to-one) multi-document configuration, in which $k$ source documents may each contribute to the target. Throughout, $n \to 1$, $1 \to n$, and $m \to n$ serve as cardinality labels; in formal statements the number of sources is~$k$.

\begin{figure}[ht!]
  \centering
  \includegraphics[width=0.95\textwidth]{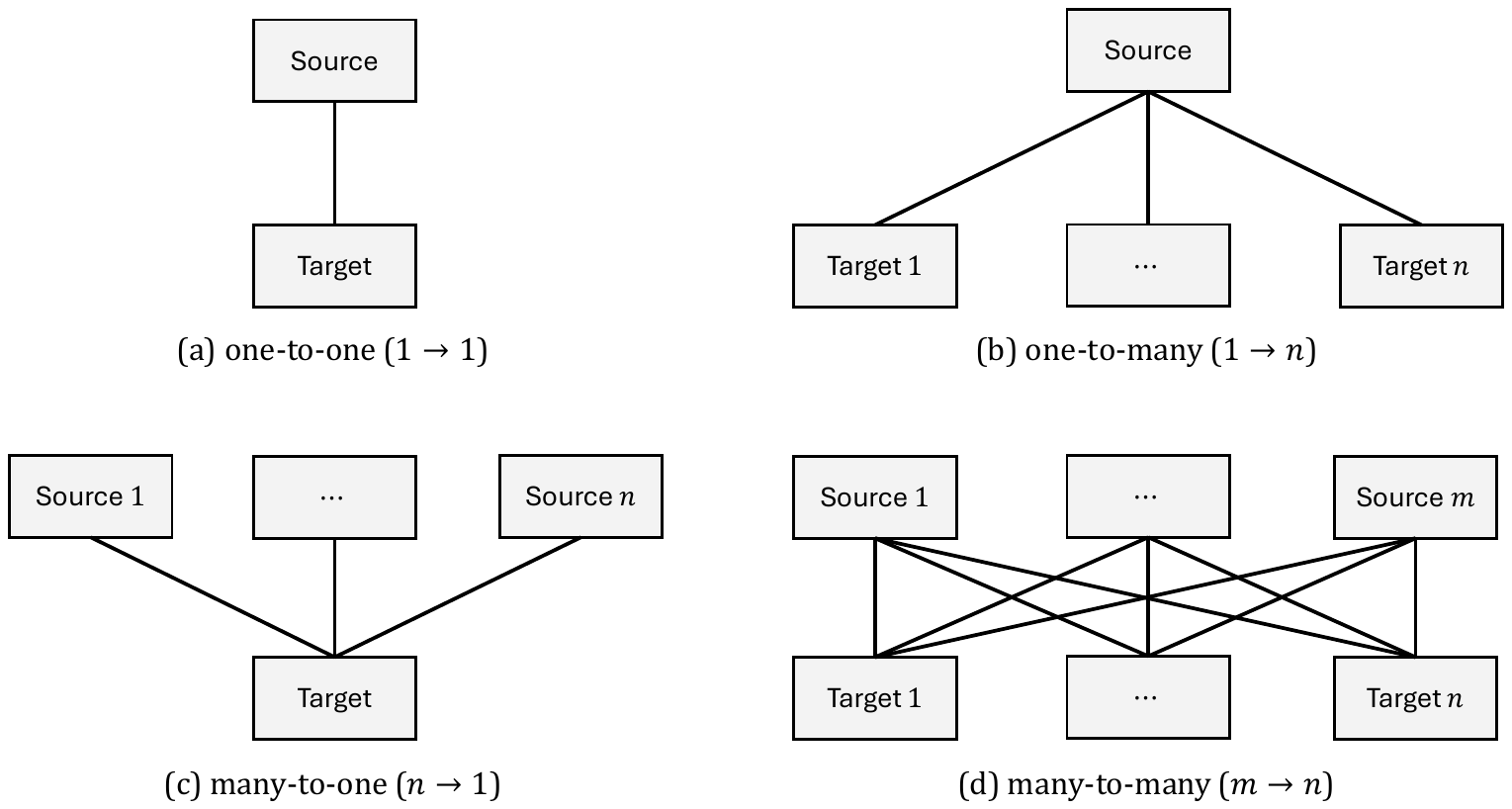}
  \caption{The four possible document cardinality configurations.
    (a)~One-to-one ($1 \to 1$): unconditioned pairwise STS.
    (b)~One-to-many ($1 \to n$): a single source with multiple targets.
    (c)~Many-to-one ($n \to 1$): the configuration developed in Sections~\ref{sec:scs}--\ref{sec:properties}.
    (d)~Many-to-many ($m \to n$): multiple sources and multiple targets.}
  \label{fig:cardinality}
\end{figure}

This work develops the framework for the $n \to 1$ case, shows that it extends to the $1 \to n$ case by vertical role reversal (Section~\ref{sec:one-to-n}), and demonstrates that the $m \to n$ case reduces to the independent composition of the two unilateral conditionings (Section~\ref{sec:m-to-n}). Together with the established $1 \to 1$ case (unconditioned pairwise STS), these configurations cover the whole document cardinality landscape illustrated in Figure~\ref{fig:cardinality}.

\subsection{Directionality of similarity relations}
\label{sec:directionality}

Classical STS treats all pairwise similarities as structurally interchangeable: $\sigma(S_i, S_j)$, $\sigma(S_i, T)$, and $\sigma(S_j, T)$ are all instances of the same undifferentiated function. In an $n \to 1$ multi-document configuration, however, the documents occupy two distinct roles---source and target---which induces a natural partition of the similarity relations into two classes.

\begin{definition}[Lateral similarity]
A similarity relation $\sigma(S_i, S_j)$ between two source documents $S_i, S_j \in \{S_1, \ldots, S_k\}$ is called a \emph{lateral} similarity. Lateral similarities measure the semantic overlap \emph{within} the source context.
\end{definition}

\begin{definition}[Vertical similarity]
A similarity relation $\sigma(S_i, T)$ between a source document $S_i$ and the target document $T$ is called a \emph{vertical} similarity. Vertical similarities measure the semantic overlap \emph{across} the source--target boundary.
\end{definition}

This distinction is not merely notational. The two classes play fundamentally different roles in the SCS framework. Lateral similarities generate the shared context (the overlaps among source documents) that must be eliminated to isolate marginal contributions. Vertical similarities are what SCS ultimately measures, after conditioning on the lateral structure.

\begin{figure}[t]
  \centering
  \includegraphics[width=0.9\textwidth]{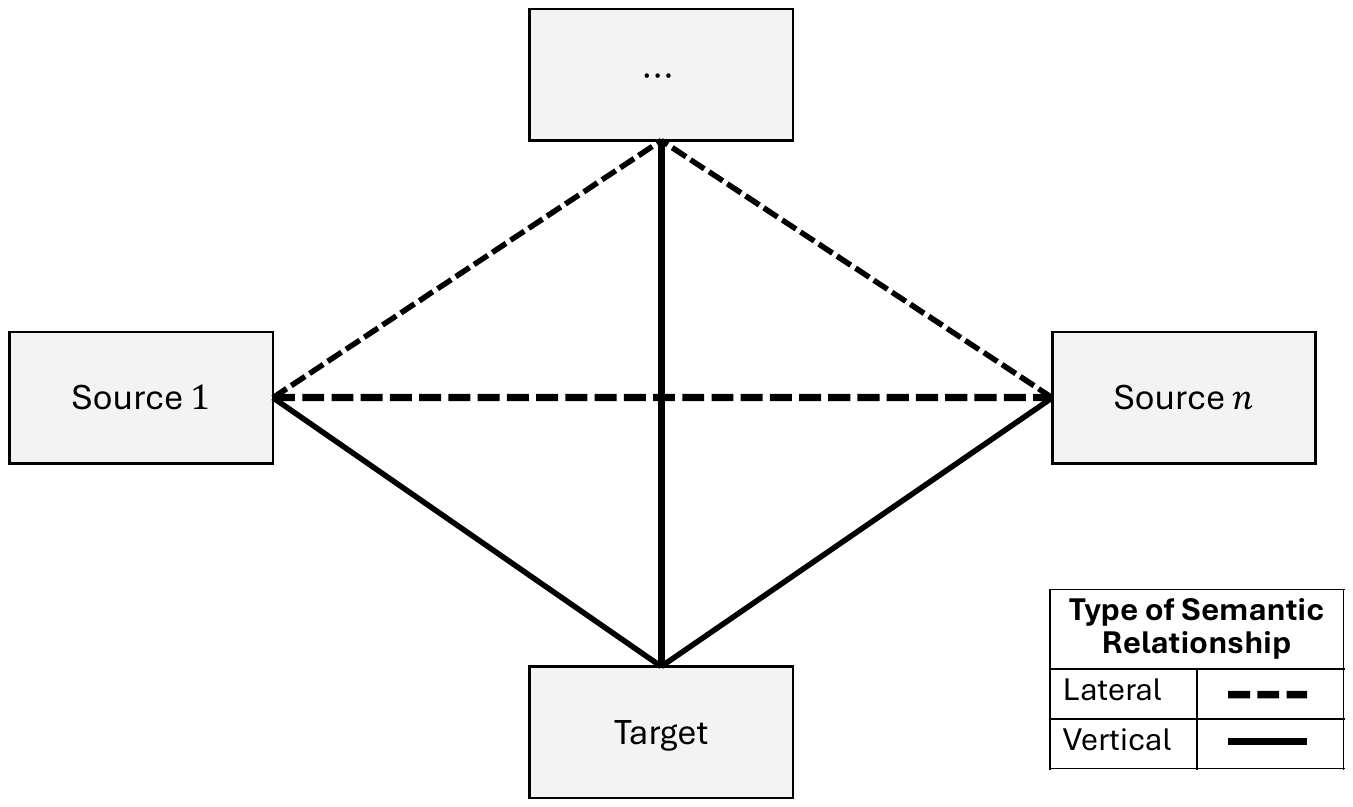}
  \caption{Directionality of similarity relations in the $n \to 1$
    case.  Lateral similarity (dashed) pairwise connects source documents
    $S_i$ and $S_j$; vertical similarities (solid) connect each
    source to the target~$T$.}
  \label{fig:directionality}
\end{figure}

Figure~\ref{fig:directionality} illustrates this partition for the $n \to 1$ case. In the $2 \to 1$ case, there is one lateral pair and two vertical pairs. In the general $n \to 1$ case, there are $\binom{k}{2}$ lateral pairs and $k$ vertical pairs. The introduction of directionality partitions the similarity space of $k+1$ documents into:
\begin{align}
  \mathcal{L} &= \{\sigma(S_i, S_j) : 1 \leq i < j \leq k\}, &\quad |\mathcal{L}| &= \tbinom{k}{2} \\
  \mathcal{V} &= \{\sigma(S_i, T) : 1 \leq i \leq k\}, &\quad |\mathcal{V}| &= k
\end{align}

This partition is a departure from classical STS, in which all $\binom{k+1}{2}$ pairwise similarities belong to a single undifferentiated set.

\subsection{Zone decomposition ($2 \to 1$ case)}

For two source documents $S_1, S_2$ and one target $T$, the conceptual space decomposes into seven zones as depicted in Figure~\ref{fig:regions} and further explained in Table~\ref{tab:regions}.:

\begin{figure}[t]
	\centering
	\includegraphics[width=0.95\textwidth]{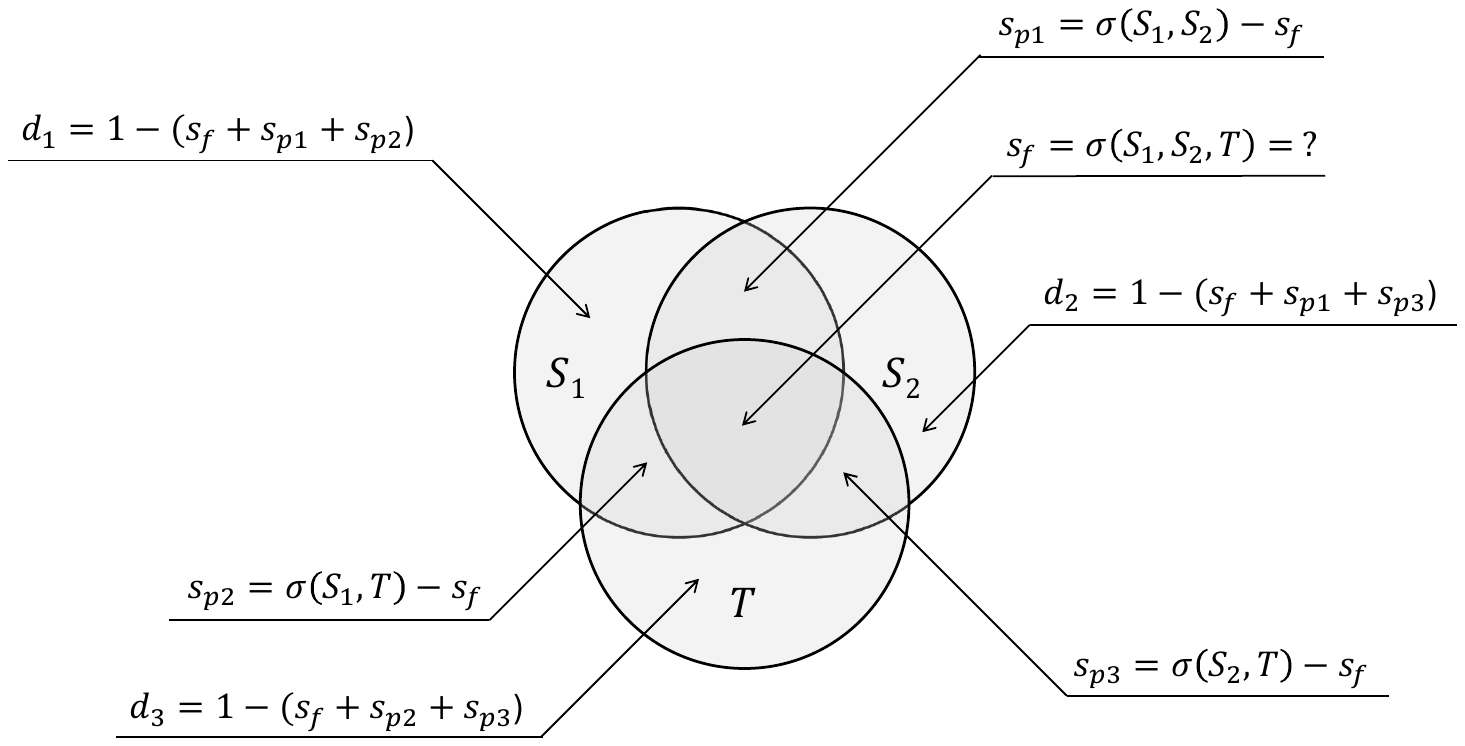}
	\caption{Three-circle Venn diagram for the $2 \to 1$ case,
		showing the seven semantic zones: full similarity~$s_f$,
		partial similarities $s_{p1}$, $s_{p2}$, $s_{p3}$, and full
		dissimilarities $d_1$, $d_2$, $d_3$.}
	\label{fig:regions}
\end{figure}

\begin{table}[h]
\centering
\begin{tabular}{lll}
\toprule
Zone & Symbol & Meaning \\
\midrule
Full similarity & $s_f$ & Shared by all three documents \\
Partial similarity 1 & $s_{p1}$ & Shared by $S_1$ and $S_2$ only (not $T$) \\
Partial similarity 2 & $s_{p2}$ & Shared by $S_1$ and $T$ only (not $S_2$) \\
Partial similarity 3 & $s_{p3}$ & Shared by $S_2$ and $T$ only (not $S_1$) \\
Full dissimilarity 1 & $d_1$ & Unique to $S_1$ \\
Full dissimilarity 2 & $d_2$ & Unique to $S_2$ \\
Full dissimilarity 3 & $d_3$ & Unique to $T$ \\
\bottomrule
\end{tabular}
\caption{Zone decomposition for the $2 \to 1$ case.}
\label{tab:regions}
\end{table}

Each document's unit-measure constraint yields:
\begin{align}
  d_1 + s_{p1} + s_{p2} + s_f &= 1 \\
  d_2 + s_{p1} + s_{p3} + s_f &= 1 \\
  d_3 + s_{p2} + s_{p3} + s_f &= 1
\end{align}

The symmetry property gives three additional equations (any one of which follows from the other two combined with the unit-measure constraints):
\begin{align}
  (d_1 + s_{p2}) - (d_2 + s_{p3}) &= 0 \\
  (d_2 + s_{p1}) - (d_3 + s_{p2}) &= 0 \\
  (d_1 + s_{p1}) - (d_3 + s_{p3}) &= 0
\end{align}

\subsection{General region counting}

\begin{theorem}[Region counting]
For $k$ source documents and one target ($k+1$ documents total), the total number of distinct semantic regions is:
\begin{equation}
  N_k = 2^{k+1} - 1
\end{equation}
\end{theorem}

This follows from the recurrence $N_k = 2\,N_{k-1} + 1$, proved by induction.

\begin{observation}
There is exactly \textbf{one} region of full semantic similarity $s_f$ (shared by all $k+1$ documents).
\end{observation}

\begin{observation}
There are exactly $k+1$ regions of full semantic dissimilarity $d_j$ (one per document).
\end{observation}

\begin{observation}
The number of partial similarity regions is $2^{k+1} - k - 3$.
\end{observation}

\subsection{Underdetermination result}

The region decomposition of Section~3.2 and its Venn diagram (Figure~\ref{fig:regions}) provide geometric intuition for the $2 \to 1$ case: three circles on a plane produce seven regions whose areas can be labeled and constrained. However, this planar representation does not extend beyond three documents---for $k+1 > 3$, no arrangement of circles in the plane can realize all $2^(k+1) - 1$ non-empty intersections. The argument must therefore be stated in terms of unit-measure regions in an $(k+1)$-dimensional space, not circles with a unit-area constraint on the plane.

\begin{theorem}[Underdetermination]
$k+1 \geq 3$ sets of unit measure with all mutual intersections non-empty produce at most $2^{k+1} - 1$ distinct zones, but only $(k+1) + \binom{k+1}{2}$ independent equations from unit-measure and pairwise-overlap constraints. The system is underdetermined whenever all zones are non-empty.
\end{theorem}

\begin{proof}
In the general-position case (all zones non-empty), the $k+1$ mutually intersecting unit-measure sets produce $2^{k+1} - 1$ distinct zones, each an unknown measure variable. The available equations are: $\binom{n}{1} = n$ unit-measure equations and $\binom{k+1}{2}$ pairwise intersection equations. We must show that $2^{k+1} - 1 > (k+1) + \binom{k+1}{2}$ for all $k+1 \geq 3$. Expanding the left side via the binomial theorem and substituting $k+1 = \binom{k+1}{1}$:
\begin{align}
  2^{k+1} - 1 &> \tbinom{k+1}{2} + (k+1) \notag\\[4pt]
  \sum_{i=1}^{k+1} \tbinom{k+1}{i} &> \tbinom{k+1}{2} + \tbinom{k+1}{1} \notag\\[4pt]
  \sum_{i=3}^{k+1} \tbinom{k+1}{i} + \tbinom{k+1}{2} + \tbinom{k+1}{1} &> \tbinom{k+1}{2} + \tbinom{k+1}{1} \notag\\[4pt]
  \sum_{i=3}^{k+1} \tbinom{k+1}{i} &> 0
\end{align}
The final inequality holds for all $k+1 \geq 3$, since every binomial coefficient $\binom{k+1}{i}$ is strictly positive.
\end{proof}

For $k+1 = 3$, this is 7~unknowns but only 6~equations. The missing degree of freedom is $s_f$---the measure of the three-way intersection, which no combination of pairwise information can resolve. More generally, the surplus unknowns correspond precisely to the higher-order intersections (triples and above), which are invisible to pairwise measurements.

Unit-measure constraints and pairwise overlaps do not determine every zone when three or more texts interact. The Venn diagram offers useful intuition for small $k$ but cannot serve as a computational foundation for arbitrary $k$. The remainder of this paper resolves this limitation by grounding document regions in concrete lemma sets, where each element either belongs to a document or does not, making every region explicit and computable regardless of the number of documents.

%------------------------------------------------------------------
\section{Departures from Classical STS}
\label{sec:departures}

The multi-document setting introduces two structural properties that distinguish SCS from classical STS. Both arise naturally from the $n \to 1$ multi-document configuration and are formalized in detail in Sections~\ref{sec:directionality} and~\ref{sec:asymmetry} respectively.

\paragraph{Departure 1: Directionality.} The lateral--vertical partition (Section~\ref{sec:directionality}) is a departure in corpus structure: SCS partitions pairwise similarities into functionally distinct classes based on document roles.

\paragraph{Departure 2: Asymmetry.} Metric-induced pairwise similarity inherits the symmetry of the underlying distance and cannot represent role-sensitive conditional comparisons. SCS enlarges the domain from unordered pairs to ordered, conditioned triples, producing role-induced asymmetry: Section~\ref{sec:asymmetry} proves that $\SCS(T, S_i \mid S_j) \neq \SCS(T, S_j \mid S_i)$ in general.

Together, these two departures allow SCS to answer a question that symmetric, undirected STS cannot: \emph{which source document contributes more uniquely to the target, and by how much?}

%------------------------------------------------------------------
\section{Transition to Set-Theoretic Representation}
\label{sec:bridge}

The \emph{principle of compositionality} states that the meaning of a complex expression is determined by the meanings of its constituents and the rules used to combine them \citep{partee1984compositionality}. The principle enables systematic interpretation of language and extends to complexity analysis across domains including mathematics, computer science, and epistemology \citep{janssen2012compositionality}.

The principle has known limits: it does not handle idiomatic expressions, metaphors, or context-dependent meanings well \citep{szabo2024compositionality}. However, these exceptions do not apply when the texts under analysis rely on standardized terminology, explicit definitions, and normative discourse---genres in which compositional clarity supports meaningful semantic decomposition.

Accordingly, each document is decomposed into its lexical elements. Documents are tokenized, lemmatized, and represented as \emph{ordered sets of lemmas}.

\begin{definition}[Preprocessing map]\label{def:preprocessing}
Let $\Sigma$ be a finite lemma universe. Define the deterministic preprocessing map
\begin{equation}
  \Lambda : \mathcal{D} \to 2^{\Sigma},
\end{equation}
where $\Lambda(d)$ is the deduplicated lemma set of document~$d$. By a standard abuse of notation, we write $S_i$ for both the $i$-th source document and its lemma set $\Lambda(S_i)$, and $T$ for both the target document and $\Lambda(T)$, from this point on.
\end{definition}

All lemma sets satisfy $S_1, \ldots, S_k, T \subseteq \Sigma$. This representation replaces geometric regions with Boolean operations on finite lemma sets. 

The preprocessing map bridges the geometric framework of Sections 2 and 3 with the set-theoretic operations that follow, collapsing the text-processing pipeline into a single deterministic step that makes the residual comparisons computable and reproducible.

\paragraph{Representation assumption.} The lemma-set representation treats lexical overlap as a proxy for semantic overlap. This is an approximation: synonymy, polysemy, and other non-compositional phenomena mean that lexical co-occurrence neither necessary nor sufficient for semantic alignment. The approximation is consistent with much of the STS literature, which has long used surface-level features as proxies for meaning; its limitations are discussed in Section~\ref{sec:discussion}.

%------------------------------------------------------------------
\section{Semantic Conditional Similarity (SCS)}
\label{sec:scs}

\subsection{Cosine Similarity on Token Sets}
\label{sec:measures}

For non-zero vectors $\vect{x}, \vect{y}$ in an inner product space:
\begin{equation}
	s_{\cos}(\vect{x}, \vect{y}) = \frac{\vect{x} \cdot \vect{y}}{\norm{\vect{x}} \, \norm{\vect{y}}}
	\label{eq:cosine}
\end{equation}
where $\vect{x}$ and $\vect{y}$ are term-frequency (or binary) vectors derived from the lemma sets, and $\norm{\cdot}$ denotes the Euclidean norm.

Cosine similarity is a similarity function, not a distance function. Its complement $1 - s_{\cos}$ fails the triangle inequality and is therefore not a metric; however, the angular distance $\tfrac{1}{\pi}\arccos(s_{\cos})$ is a true metric on the unit sphere.

\subsection{Key insight}

\begin{figure}[H]
	\centering
	\includegraphics[width=0.95\textwidth]{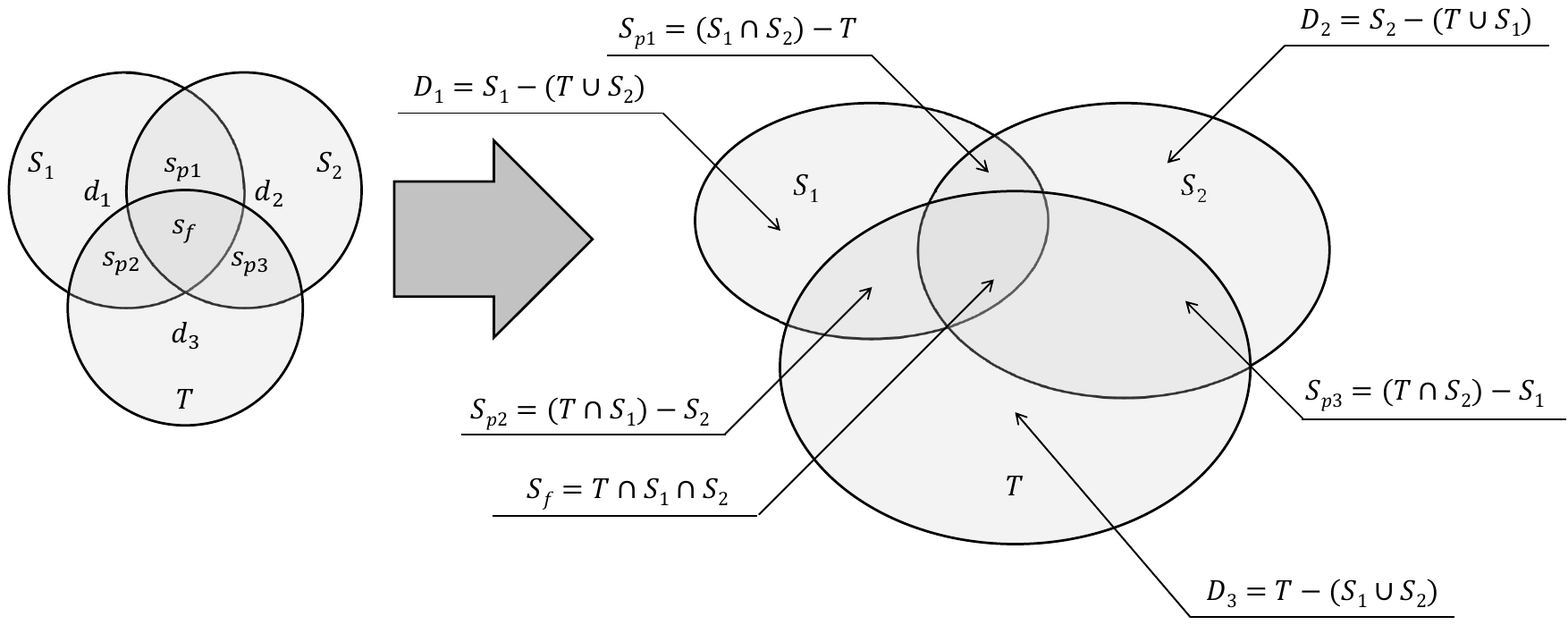}
	\caption{From geometric zones to residual sets.  The seven zones of
		the Venn diagram (Figure~\ref{fig:regions}) are mapped to set-theoretic
		expressions.  The lateral shared context $I_{RC} = R \cap C$ is
		subtracted from both reference and target, yielding the reference
		residual~$U_{R|C}$ and the target residual~$T \setminus (R \cap C)$.  The SCS
		score is computed on these residuals.}
	\label{fig:setops}
\end{figure}

The intersection $S_1 \cap S_2$ spans the zones $s_{p1}$ 
and $s_f$ of Figure~\ref{fig:regions} and is not a 
differentiating factor for the vertical semantic relationship between either $S_1$ and $T$ or $S_2$ and $T$. Instead, one must isolate the \emph{marginal (residual) semantic contribution} of each source document by conditioning on the other as shown in  Figure~\ref{fig:setops}. Table~\ref{tab:regionmap} further tabulates the set expressions for each geometric zone shown in the figure.

\subsection{Definition}

\begin{definition}[Semantic Conditional Similarity]
	\emph{Semantic Conditional Similarity (SCS)} is a conditional form of Semantic Textual Similarity in which the similarity between two documents is computed after eliminating semantic elements shared with another document in a given context. SCS captures the marginal (net or residual) semantic contribution of one document to another by conditioning on pairs of documents under analysis within the specified context.
\end{definition}

\subsection{Notation}

For $k$ source documents with lemma sets $S_1, \ldots, S_k$ and target with lemma set $T$, each conditioning operation selects one source as the \emph{reference document}~$R$ (the source whose marginal contribution is being measured) and another as the \emph{conditioning document}~$C$ (the source whose shared content is subtracted). For an ordered pair $(i, j)$ with $i \neq j$, we set $R = S_i$ and $C = S_j$ and write
\begin{equation}
	\SCS(T, R \mid C)
\end{equation}
for the \emph{marginal semantic similarity between the target and the reference document, conditional on the conditioning document}.

\vfill

\begin{table}[H]
	\centering
	\begin{tabular}{ll}
		\toprule
		Zone & Set expression \\
		\midrule
		$s_{p1}$ & $(S_1 \cap S_2) - T$ \\
		$s_{p2}$ & $(T \cap S_1) - S_2$ \\
		$s_{p3}$ & $(T \cap S_2) - S_1$ \\
		$s_f$ & $T \cap S_1 \cap S_2$ \\
		$d_1$ & $S_1 - (T \cup S_2)$ \\
		$d_2$ & $S_2 - (T \cup S_1)$ \\
		$d_3$ & $T - (S_1 \cup S_2)$ \\
		\bottomrule
	\end{tabular}
		\caption{Set-theoretic zone mapping for the $2 \to 1$ case.}
		\label{tab:regionmap}
\end{table}

\subsection{Directional comparison state}

The \emph{directional comparison state} is the triple $(T, R \mid C)$, read as ``target~$T$, reference~$R$, conditioned on~$C$.'' The triple encodes which source is the reference and which serves as the conditioning document---swapping $R$ and $C$ changes the comparison state and can change the score.

\subsection{Named residual sets}

\begin{definition}[Named residual sets]\label{def:residuals}
For a conditioning triplet $(T, R \mid C)$, define:
\begin{align}
  I_{RC} &= R \cap C & &\text{(lateral shared context)} \\
  U_{R|C} &= R \setminus I_{RC} = R \setminus C & &\text{(reference residual)}
\end{align}
The target enters the score computation through the expression $T \setminus (R \cap C)$: the target with the lateral shared context removed. This expression is written inline throughout, since it is self-documenting and requires no separate symbol.
\end{definition}

\subsection{Cosine-SCS formula}

\begin{definition}[Cosine-based Semantic Conditional Similarity]\label{def:cosine-scs}
Each lemma set is represented as a binary indicator vector over the 
vocabulary, where coordinate~$\mu$ is~$1$ if and only if the 
$\mu$-th lemma belongs to the set. Since both vectors are binary, 
$\langle \chi(A), \chi(B) \rangle = \sum_{\mu} \chi_A(\mu)\,\chi_B(\mu)$ 
counts exactly the positions where both are~$1$, i.e.\ $\card{A \cap B}$; likewise, the Euclidean norm $\norm{\chi(A)} = \sqrt{\card{A}}$ \citep[§\,6.3]{Manning2008}. Cosine similarity therefore reduces to set-cardinality expressions. The \emph{cosine-based SCS} is therefore
\begin{equation}
  \SCS_{\cos}(T, R \mid C)
    = \frac{\card{U_{R|C} \cap \bigl(T \setminus (R \cap C)\bigr)}}{\sqrt{\card{U_{R|C}} \cdot \card{T \setminus (R \cap C)}}},
  \label{eq:cosine-scs-set}
\end{equation}
with admissible domain $U_{R|C} \neq \emptyset$ and $T \setminus (R \cap C) \neq \emptyset$.
\end{definition}

The conditioning framework is measure-agnostic: any set-based similarity function---Jaccard, S\o{}rensen-Dice, or others---can replace cosine in the construction above. This paper develops SCS with cosine throughout, as cosine 
similarity is the dominant measure in semantic similarity 
research \citep{agirre2012semeval}. 

\subsection{Overlap identity}

\begin{theorem}[Overlap identity]
The overlap actually counted by SCS is exactly the target content aligned with $R$ but not with $C$:
\begin{equation}
  U_{R|C} \cap \bigl(T \setminus (R \cap C)\bigr) = T \cap (R \setminus C).
\end{equation}
\end{theorem}

\begin{proof}
\begin{align}
  U_{R|C} \cap \bigl(T \setminus (R \cap C)\bigr)
    &= (R \setminus C) \cap \bigl(T \setminus (R \cap C)\bigr) \notag\\
    &= R \cap C^c \cap T \cap (R^c \cup C^c) \notag\\
    &= T \cap R \cap C^c \notag\\
    &= T \cap (R \setminus C). \qedhere
\end{align}
\end{proof}

The score counts target content aligned with the reference document after removal of the content shared with the conditioning document.

\subsection{Role-induced asymmetry}
\label{sec:asymmetry}

As noted in Section~\ref{sec:departures}, extending the comparison domain to ordered triples induces asymmetric scores: the underlying cosine function remains symmetric, but swapping the reference and conditioning roles changes its inputs. We now prove this formally.

\begin{theorem}[Asymmetry of SCS]
SCS is not symmetric in general: swapping the reference and conditioning roles can change the score.
\end{theorem}

\begin{proof}[Proof by counterexample]
Let $i \neq j$ with $S_i = \{a, b, c\}$, $S_j = \{c, d\}$, $T = \{a, b, d\}$.

Setting $R = S_i$ and $C = S_j$:
\begin{align}
  I_{RC} &= S_i \cap S_j = \{c\}, \notag\\
  U_{R|C} &= S_i \setminus S_j = \{a, b\}, \quad T \setminus (R \cap C) = T \setminus \{c\} = \{a, b, d\}, \notag
\end{align}
so $U_{R|C} \cap \bigl(T \setminus (R \cap C)\bigr) = \{a, b\}$ and
\begin{equation}
  \SCS_{\cos}(T, S_i \mid S_j) = \frac{2}{\sqrt{2 \cdot 3}} = \frac{2}{\sqrt{6}}.
\end{equation}

Now swap the roles: set $R = S_j$ and $C = S_i$. Then
\begin{align}
  I_{RC} &= S_j \cap S_i = \{c\}, \notag\\
  U_{R|C} &= S_j \setminus S_i = \{d\}, \quad T \setminus (R \cap C) = T \setminus \{c\} = \{a, b, d\}, \notag
\end{align}
so $U_{R|C} \cap \bigl(T \setminus (R \cap C)\bigr) = \{d\}$ and
\begin{equation}
  \SCS_{\cos}(T, S_j \mid S_i) = \frac{1}{\sqrt{1 \cdot 3}} = \frac{1}{\sqrt{3}}.
\end{equation}
Therefore $\SCS_{\cos}(T, S_i \mid S_j) = 2/\sqrt{6} \neq 1/\sqrt{3} = \SCS_{\cos}(T, S_j \mid S_i)$.
\end{proof}

The SCS score matrix for $k$ source documents is therefore a \emph{non-symmetric} $k \times k$ matrix (with undefined diagonal), where entry $(i, j) = \SCS_{\cos}(T, S_i \mid S_j)$. This contrasts with the symmetric similarity matrix of classical STS. The implications of this asymmetry for similarity theory are discussed in Section~\ref{sec:discussion}.

\subsection{SCS score count}

For $k$ source documents, the procedure is repeated across all ordered source pairs $(i, j)$ with $i \neq j$. This yields $k \times (k-1)$ SCS scores. For $k = 3$, this gives 6 scores.

\subsection{Admissibility and averaging}

For a collection of source texts compared with a common target, define the \emph{admissible conditioning indices} for source~$i$:
\begin{equation}
  J_i = \{j \neq i : U_{R|C} \neq \emptyset \text{ and } T \setminus (R \cap C) \neq \emptyset\},
\end{equation}
where $R = S_i$ and $C = S_j$.

The \emph{average conditional score} for source~$i$ with respect to target~$T$ is then:
\begin{equation}
  \overline{\SCS}_{\cos}(T, S_i)
    = \frac{1}{\card{J_i}} \sum_{j \in J_i} \SCS_{\cos}(T, S_i \mid S_j).
\end{equation}

The average can support ranking sources within a specified comparison design. Note that the average depends on the comparison set: adding or removing a source document changes the conditioning partners available and may therefore change the average scores of the existing sources. A causal or historical interpretation requires additional evidence beyond the similarity score.

%------------------------------------------------------------------
\subsection{Algorithmic Workflow}
\label{sec:algorithm}

The SCS algorithm proceeds as follows: 

\medskip\noindent\textbf{Input:} $k$ source documents $\{S_1, \ldots, S_k\}$, one target document $T$.

\begin{enumerate}
  \item \textbf{Preprocessing.} Each document undergoes tokenization into lemma lists (excluding stop words, punctuation, numerical tokens).
  \item \textbf{Set construction.} Convert lemma lists into ordered lemma sets $\{S_1, \ldots, S_k, T\}$, eliminating duplicate lemmas.
  \item \textbf{Pairwise decomposition.} For each source pair $(S_i, S_j)$, compute:
    \begin{itemize}
      \item Intersection: $S_i \cap S_j$
      \item Complement of $S_i$ given $S_j$: $S_i - (S_i \cap S_j)$
      \item Complement of $S_j$ given $S_i$: $S_j - (S_i \cap S_j)$
      \item Conditioned target: $T - (S_i \cap S_j)$
    \end{itemize}
  \item \textbf{Vectorization.} Transform all complement sets into binary document vectors over a common vocabulary (union of all lemmas). Each dimension corresponds to a lemma; presence~$= 1$, absence~$= 0$.
  \item \textbf{Similarity computation.} Compute cosine similarity between each conditioned source vector and the corresponding conditioned target vector to obtain SCS scores $\SCS_{\cos}(T, R \mid C)$, where $R = S_i$ and $C = S_j$.
  \item \textbf{Ranking.} Sort all SCS scores in descending order to identify dominant source documents.
\end{enumerate}

\noindent\textbf{Output:} A ranked profile of $k \times (k-1)$ SCS scores, revealing the marginal semantic contribution of each source document to the target.

\vspace{1em}

\begin{figure}[t]
  \centering
  \includegraphics[width=\textwidth]{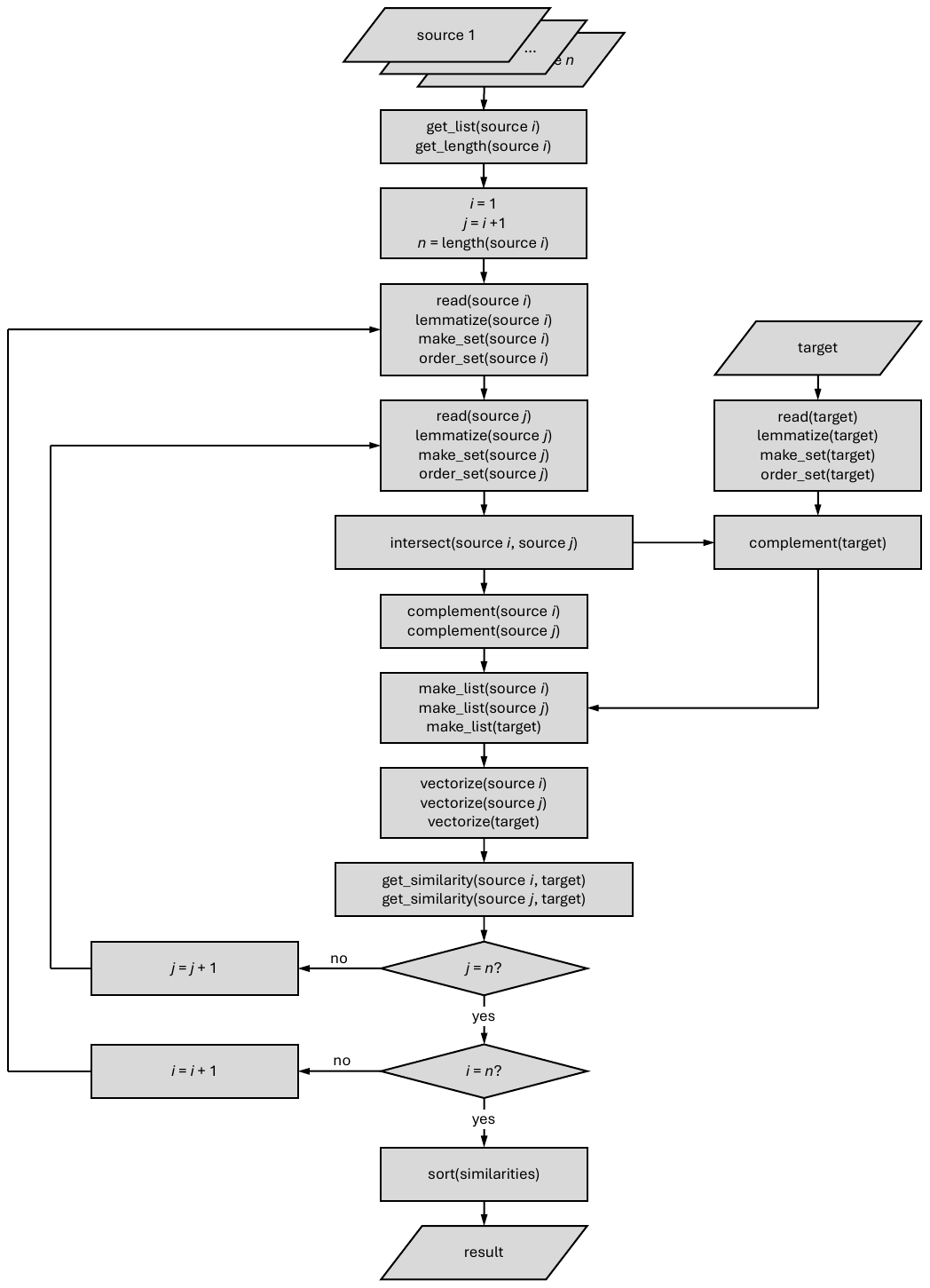}
  \caption{Computational pipeline for Semantic Conditional Similarity (SCS).
    Source documents are read, lemmatized, and converted to ordered sets
    (Definition~\ref{def:preprocessing}).  For each ordered source pair $(i, j)$, setting $R = S_i$
    and $C = S_j$, intersection and complement operations produce the residual
    sets $U_{R|C}$ and $T \setminus (R \cap C)$ (Definition~\ref{def:residuals}).  Cosine similarity
    on the binary indicator vectors of these residuals reduces to set-cardinality
    expressions (Equation~\ref{eq:cosine-scs-set}).  The nested loop over
    all ordered pairs yields $k \times (k{-}1)$ scores, which are sorted to
    produce the final ranked profile.}
  \label{fig:scs-flowchart}
\end{figure}

Figure~\ref{fig:scs-flowchart} provides a visual overview of the computational pipeline, mapping each stage to the formal definitions of Sections~\ref{sec:bridge}--\ref{sec:scs}.

\paragraph{Complexity analysis} Let $k$ be the number of sources, $L_d$ the document length before lemmatization, $\ell$ the average resulting lemma-set size, and $v$ the active vocabulary size.

\paragraph{Preprocessing.} Lemmatization and deduplication run in $O\!\bigl(\sum_d L_d\bigr)$, where $O(\cdot)$ denotes asymptotic upper-bound complexity.

\paragraph{Pairwise loop.} The pair loop visits $\binom{k}{2}$ unordered source pairs and computes two directional scores per pair. With hash-set operations and explicit vectorization, the dominant cost is
\begin{equation}
  O\!\bigl(\tbinom{k}{2} \cdot (\ell + v)\bigr),
\end{equation}
plus $O(k \log k)$ for sorting the averaged scores.

Direct set-cardinality computation can avoid explicit vectorization, since cosine on binary indicator vectors reduces to set-cardinality expressions (Definition~\ref{def:cosine-scs}).

%------------------------------------------------------------------
\section{Properties and Constraints}
\label{sec:properties}

\subsection{Formal properties}

\begin{theorem}[Boundedness]
For every admissible triple,
\begin{equation}
  0 \leq \SCS_{\cos}(T, R \mid C) \leq 1.
\end{equation}
\end{theorem}

\begin{proof}
The cosine of nonzero binary vectors lies in $[0, 1]$.
\end{proof}

\begin{theorem}[Collapse to STS]
If $R \cap C = \emptyset$, then SCS collapses to ordinary set-based cosine similarity:
\begin{equation}
  \SCS_{\cos}(T, R \mid C)
    = \frac{\card{R \cap T}}{\sqrt{\card{R} \cdot \card{T}}}.
\end{equation}
\end{theorem}

\begin{proof}
If $R \cap C = \emptyset$, then $U_{R|C} = R$ and $T \setminus (R \cap C) = T$. Substituting into the cosine-SCS definition yields the standard cosine similarity formula.
\end{proof}

Together with the admissibility condition $U_{R|C} \neq \emptyset$, the Collapse theorem identifies one end of a spectrum: as the overlap between reference and conditioning document decreases toward zero, SCS approaches unconditional similarity. The opposite boundary---total overlap---is the subject of the next theorem.

\begin{theorem}[Identical sources]
\label{thm:identical-sources}
If $R = C$, then $U_{R|C} = \emptyset$ and $\SCS_{\cos}(T, R \mid C)$ is undefined.
\end{theorem}

\begin{proof}
$U_{R|C} = R \setminus (R \cap C) = R \setminus R = \emptyset$. The reference residual is the zero vector, so the cosine $\tfrac{\vect{u} \cdot \vect{v}}{\norm{\vect{u}}\norm{\vect{v}}}$ has a zero denominator.
\end{proof}

This is the correct behavior: when the conditioning document is identical to the reference, the reference has no unique content, and the question ``how much of $R$'s unique content appears in $T$?''\ is vacuous. Operationally, any pair $(R, C)$ with $R = C$ falls outside the admissible domain and is excluded from the comparison set $J_i$ (Section~\ref{sec:algorithm}).

\begin{theorem}[Identity]
Identity behaves differently on residualized objects and on raw documents.

First, on residualized objects:
\begin{equation}
  U_{R|C} = T \setminus (R \cap C) \neq \emptyset \;\Longrightarrow\;
  \SCS_{\cos}(T, R \mid C) = 1.
\end{equation}

Second, on raw documents, identity of indiscernibles fails: there exist $R \neq T$ such that $\SCS_{\cos}(T, R \mid C) = 1$.
\end{theorem}

\begin{proof}
The first claim is immediate because any set is maximally similar to itself. For the second claim, take $R = \{a, b, c\}$, $C = \{b, c\}$, $T = \{a, b\}$. Then $U_{R|C} = R \setminus C = \{a\}$, $T \setminus (R \cap C) = T \setminus \{b, c\} = \{a\}$. So cosine-SCS equals~$1$, even though $R \neq T$.
\end{proof}

The conditional score is not a metric on raw texts. Identity on residualized objects does not imply identity of raw documents.

\begin{theorem}[Monotonicity]
SCS has the expected conditional monotonicity properties.

If $U \subseteq T$ and $M \subseteq T \setminus U$, then adding \emph{matched} source material increases similarity:
\begin{equation}
  \cos(U \cup M, T) \geq \cos(U, T).
\end{equation}

If $N \cap T = \emptyset$, then adding \emph{unmatched} source material decreases similarity:
\begin{equation}
  \cos(U \cup N, T) \leq \cos(U, T).
\end{equation}

Likewise, if $N \cap U = \emptyset$, then adding unmatched target material decreases similarity:
\begin{equation}
  \cos(U, T \cup N) \leq \cos(U, T).
\end{equation}
\end{theorem}

\begin{proof}
If $U \subseteq T$, then $\cos(U, T) = \card{U}/\sqrt{\card{U} \cdot \card{T}} = \sqrt{\card{U}/\card{T}}$, which increases with $\card{U}$. When unmatched material is added, the intersection size stays fixed while the denominator grows, so cosine weakly decreases.
\end{proof}

Matched source and target material can raise the score; unmatched additions increase a denominator without increasing its numerator.

\subsection{Structural constraints}

\begin{itemize}
  \item SCS is defined for $k \geq 2$ source documents (the conditioning requires at least one other source document).
  \item The framework assumes a \emph{single target} (the $n \to 1$ constraint).
  \item Binary document vectors ensure that lexical frequency does not distort the conditional analysis; only the presence or absence of lemmas matters.
  \item The method is \emph{scalable} in the number of source documents, with $O(k^2)$ pairwise comparisons (Section~\ref{sec:algorithm}).
 \item SCS introduces \emph{directionality}  (Section~\ref{sec:directionality}), using lateral similarities to  condition vertical measurements.
 \item SCS exhibits \emph{role-induced asymmetry}  (Section~\ref{sec:asymmetry}), producing a non-symmetric score matrix  that distinguishes the direction of marginal contribution.
\end{itemize}

\subsection{Ordinal interpretation}

SCS scores should be interpreted \emph{ordinally}, not absolutely. The algorithm's final operation is sorting: it produces a ranked profile of conditional pairwise similarities, not a set of absolute values. This is an epistemological constraint, not a technical limitation. Semantics is not a precise science---the cosine similarities computed over binary document vectors are quantitative approximations of inherently imprecise semantic relationships. While the set-theoretic decomposition stage is exact (a lemma is either in an intersection or it is not), the similarity scores derived from those decomposed sets are best treated as providing an \emph{ordering} of marginal contributions rather than as exact magnitudes. SCS tells which source document contributes more uniquely to the target, and by roughly how much relative to the others---but the individual scores should not be taken as precise measurements of semantic content.

The appropriate use of SCS output is therefore to \emph{rank} conditional pairwise document similarities, identifying dominant source documents. When the number of source documents $k$ is sufficiently large, the $k \times (k-1)$ scores may also support clustering to identify groups of source documents with similar marginal contribution profiles. Since each source document generates $k-1$ scores (one per conditioning partner), the feasibility of clustering depends on whether $k-1$ is large enough to constitute a meaningful profile that distinguishes genuine patterns from noise. For small $k$, ranking is the only viable interpretive operation.

%------------------------------------------------------------------
\section{Extensions Across the Document Cardinality Spectrum}
\label{sec:extensions}

Sections~\ref{sec:scs}--\ref{sec:properties} developed SCS for the $n \to 1$ (many-to-one) configuration. This section shows that the framework extends naturally to the remaining document cardinalities, completing the spectrum introduced in Section~\ref{sec:decomposition}. Figure~\ref{fig:cardinality} illustrates the four configurations.

\subsection{The $1 \to 1$ case: unconditioned pairwise STS}
\label{sec:one-to-one}

The $1 \to 1$ configuration is the established baseline: a single source document $S$ and a single target document $T$. With only one source, there is no second source to serve as a conditioning document, and the SCS conditioning operation is vacuous. The similarity reduces to unconditioned pairwise STS:
\begin{equation}
	\sigma(S, T) = \frac{\card{S \cap T}}{\sqrt{\card{S} \cdot \card{T}}}.
\end{equation}
This is precisely the Collapse theorem (Section~\ref{sec:properties}): when $R \cap C = \emptyset$---or, equivalently, when no conditioning document exists---SCS reduces to ordinary cosine similarity. The $1 \to 1$ case therefore sits at the unconditioned end of the spectrum, and every standard STS result applies without modification.

\subsection{The $n \to 1$ case: source-side conditioning}
\label{sec:n-to-one-recap}

The $n \to 1$ configuration is the primary setting of this paper: $k$ source documents $\{S_1, \ldots, S_k\}$ may each have contributed to a single target~$T$. The analytical question is how much each source contributed uniquely, net of what it shares with the other sources. SCS addresses this by conditioning on the lateral overlap among sources, producing $k \times (k-1)$ directional scores. The full development appears in Sections~\ref{sec:scs}--\ref{sec:properties}.

\subsection{Extension to the $1 \to n$ configuration}
\label{sec:one-to-n}

Although this paper develops SCS for the $n \to 1$ (many-to-one) configuration, the framework extends naturally to the $1 \to n$ (one-to-many) case through a vertical role reversal of the conditioning structure. The key observation is that the confound addressed by SCS is symmetric across the source--target boundary: in the $n \to 1$ case, source documents share vocabulary with one another, inflating every source's unconditioned similarity to the target; in the $1 \to n$ case, target documents share vocabulary with one another, inflating the single source's unconditioned similarity to every target.

In the $1 \to n$ configuration, a single source document $S$ has a semantic relationship with multiple target documents $T_1, T_2, \ldots, T_n$. The analytical question is no longer ``how much did each source uniquely contribute to the target?'' but rather ``how much of the source's content did each target uniquely receive?'' The roles transpose as follows:

\begin{itemize}
	\item The \textbf{source} $S$ is fixed, replacing the fixed target of the $n \to 1$ case.
	\item \textbf{Lateral} similarities are now among targets: $\sigma(T_i, T_j)$. These generate the shared context that must be eliminated.
	\item \textbf{Vertical} similarities remain source-to-target: $\sigma(S, T_i)$. These are what SCS ultimately measures after conditioning.
	\item Each conditioning operation selects one target as the \emph{reference target} $T_i$ (the target whose unique reception of $S$ is being measured) and another as the \emph{conditioning target} $T_j$ (the target whose shared content is subtracted).
\end{itemize}

The residual sets for the $1 \to n$ case are:
\begin{align}
	I_{T_iT_j} &= T_i \cap T_j & &\text{(lateral shared context among targets)} \\
	U_{T_i|T_j} &= T_i \setminus T_j & &\text{(reference target residual)} \\
	U_{S|T_iT_j} &= S \setminus (T_i \cap T_j) & &\text{(source residual)}
\end{align}
and the $1 \to n$ SCS score is
\begin{equation}
	\SCS_{\cos}(S, T_i \mid T_j)
	= \frac{\card{U_{T_i|T_j} \cap \bigl(S \setminus (T_i \cap T_j)\bigr)}}
	{\sqrt{\card{U_{T_i|T_j}} \cdot \card{S \setminus (T_i \cap T_j)}}}.
\end{equation}

All four contributions of SCS---directionality, asymmetry, inter-document conditioning, and threshold-free partitioning---carry over intact, with the lateral structure now lying among targets rather than among sources.

The $1 \to n$ case has natural applications wherever a single source text has diffused into multiple receiving documents. For example, if a constitutional principle has influenced several statutes, the $1 \to n$ SCS would identify which statute received the most distinctive content from that principle, net of what the statutes share with one another.

\subsection{Extension to the $m \to n$ configuration}
\label{sec:m-to-n}

The $n \to 1$ and $1 \to n$ frameworks condition on one side of the source--target boundary: the former eliminates lateral overlap among sources, the latter among targets. The $m \to n$ (many-to-many) configuration, in which $m$ sources and $n$ targets interact simultaneously, requires conditioning on \emph{both} sides. The key observation is that the two conditioning operations are \emph{independent}: source-side conditioning acts only on source documents, target-side conditioning acts only on target documents, and the two document sets are disjoint. Because the operations do not interact, they can be composed without ordering effects.

For a specific source--target pair $(S_i, T_j)$ in an $m \to n$ corpus with source set $\{S_1, \ldots, S_m\}$ and target set $\{T_1, \ldots, T_n\}$:
\begin{enumerate}
	\item Compute the source-side residual $U_{S_i|S_{-i}} = S_i \setminus \bigcup_{h \neq i}(S_i \cap S_h)$, as in the $n \to 1$ framework.
	\item Compute the target-side residual $U_{T_j|T_{-j}} = T_j \setminus \bigcup_{l \neq j}(T_j \cap T_l)$, as in the $1 \to n$ framework.
	\item The \emph{bilaterally conditioned} SCS is
\end{enumerate}
\begin{equation}
	\SCS_{\cos}(S_i, T_j \mid S_{-i}, T_{-j})
	= \frac{\card{U_{S_i|S_{-i}} \cap U_{T_j|T_{-j}}}}
	{\sqrt{\card{U_{S_i|S_{-i}}} \cdot \card{U_{T_j|T_{-j}}}}}.
\end{equation}

This measures the cosine similarity between the content that is unique to source~$i$ among all sources and the content that is unique to target~$j$ among all targets. The numerator captures the lemmas that appear in exactly one source and exactly one target---the \emph{exclusive bilateral} overlap.

The bilateral decomposition yields three complementary $m \times n$ matrices for any corpus:
\begin{itemize}
	\item $M_S[i,j] = \SCS_{\cos}(T_j, S_i \mid S_{-i})$: source-conditioned (the $n \to 1$ view), measuring how much of each source's unique content aligns with each target.
	\item $M_T[i,j] = \SCS_{\cos}(S_i, T_j \mid T_{-j})$: target-conditioned (the $1 \to n$ view), measuring how much of each target's unique content aligns with each source.
	\item $M_B[i,j] = \SCS_{\cos}(S_i, T_j \mid S_{-i}, T_{-j})$: bilaterally conditioned, measuring the overlap between source-unique and target-unique content.
\end{itemize}
Comparing the three matrices entry by entry reveals how each side of the boundary contributes to or attenuates the observed similarity. A high $M_S[i,j]$ paired with a low $M_B[i,j]$ indicates that what source~$i$ uniquely contributes to target~$j$ is content that target~$j$ shares with other targets---the uniqueness is one-sided. Bilateral conditioning is more restrictive than either unilateral case: both residuals are subsets of the original documents, so $M_B$ entries will generally be lower and, for some pairs, may approach zero. This is a feature of the measure: it identifies genuinely exclusive source--target relationships, filtering out content that is laterally shared on either side.

All four contributions carry over to the bilateral case, with lateral and vertical classes defined on both sides of the source--target boundary simultaneously.

\subsection{Completeness of the cardinality spectrum}

Together with the $1 \to 1$ case---which is the established, unconditioned pairwise STS---the $n \to 1$, $1 \to n$, and $m \to n$ configurations exhaust the document cardinality landscape. SCS thus provides a complete family of conditional similarity measures for all multi-document configurations, unified by the same set-theoretic conditioning mechanism and preserving all four contributions across the spectrum.

%------------------------------------------------------------------
\section{Synthetic Demonstration}
\label{sec:demonstration}

This section provides a self-contained numerical demonstration of SCS on synthetic data. The demonstration uses single-letter lemmas to make every set operation visible, and is fully reproducible from the accompanying Python script (seed 5663).

\subsection{Setup}

Three source documents and one target are represented as ordered lemma sets over the alphabet $\{a, \ldots, z\}$:
\begin{align}
  S_1 &= \{a, b, e, f, k, m, o, q, w, y\}, & |S_1| &= 10 \notag\\
  S_2 &= \{a, f, m, p, q, t, w, y\}, & |S_2| &= 8 \notag\\
  S_3 &= \{a, c, i, j, l, o, p, t\}, & |S_3| &= 8 \notag\\
  T   &= \{b, c, d, h, i, m, p, w\}, & |T|   &= 8 \notag
\end{align}

\subsection{Unconditioned pairwise STS}

The standard cosine similarities between each source and the target are:
\begin{align}
  \sigma(S_1, T) &= \frac{|S_1 \cap T|}{\sqrt{|S_1| \cdot |T|}} = \frac{|\{b, m, w\}|}{\sqrt{10 \cdot 8}} = \frac{3}{\sqrt{80}} \approx 0.335 \notag\\[4pt]
  \sigma(S_2, T) &= \frac{|S_2 \cap T|}{\sqrt{|S_2| \cdot |T|}} = \frac{|\{m, p, w\}|}{\sqrt{8 \cdot 8}} = \frac{3}{8} = 0.375 \notag\\[4pt]
  \sigma(S_3, T) &= \frac{|S_3 \cap T|}{\sqrt{|S_3| \cdot |T|}} = \frac{|\{c, i, p\}|}{\sqrt{8 \cdot 8}} = \frac{3}{8} = 0.375 \notag
\end{align}
The unconditioned ranking is $S_2 \geq S_3 > S_1$. On pairwise STS alone, $S_2$ and $S_3$ contribute equally to the target.

\subsection{Lateral structure}

The inter-source overlaps reveal a confound:
\begin{align}
  S_1 \cap S_2 &= \{a, f, m, q, w, y\}, & |S_1 \cap S_2| &= 6 \notag\\
  S_1 \cap S_3 &= \{a, o\}, & |S_1 \cap S_3| &= 2 \notag\\
  S_2 \cap S_3 &= \{a, p, t\}, & |S_2 \cap S_3| &= 3 \notag
\end{align}
Sources $S_1$ and $S_2$ share six of their eight-to-ten lemmas. Of the three lemmas in $S_2 \cap T = \{m, p, w\}$, two ($m$ and $w$) also belong to $S_1$. The unconditioned score $\sigma(S_2, T)$ therefore counts content that $S_2$ shares with $S_1$, conflating shared lateral vocabulary with genuinely unique contribution.

\subsection{SCS scores}

Table~\ref{tab:scs-demo} reports all six SCS scores for the $3 \to 1$ configuration, computed via Equation~\ref{eq:cosine-scs-set}. Each row shows the lateral shared context $I_{RC}$, the reference residual $U_{R|C}$, the target residual $T \setminus I_{RC}$, their overlap, and the resulting score.

\begin{table}[ht!]
\centering
\small
\begin{tabular}{@{}llllcc@{}}
\toprule
SCS$(T,R \mid C)$ & $I_{RC}$ & $U_{R|C}$ & $T \setminus I_{RC}$ & Overlap & Score \\
\midrule
$\SCS(T, S_1 \mid S_2)$ & $\{a,f,m,q,w,y\}$ & $\{b,e,k,o\}$ & $\{b,c,d,h,i,p\}$ & $\{b\}$ & $0.204$ \\[2pt]
$\SCS(T, S_1 \mid S_3)$ & $\{a,o\}$ & $\{b,e,f,k,m,q,w,y\}$ & $\{b,c,d,h,i,m,p,w\}$ & $\{b,m,w\}$ & $0.375$ \\[2pt]
$\SCS(T, S_2 \mid S_1)$ & $\{a,f,m,q,w,y\}$ & $\{p,t\}$ & $\{b,c,d,h,i,p\}$ & $\{p\}$ & $0.289$ \\[2pt]
$\SCS(T, S_2 \mid S_3)$ & $\{a,p,t\}$ & $\{f,m,q,w,y\}$ & $\{b,c,d,h,i,m,w\}$ & $\{m,w\}$ & $0.338$ \\[2pt]
$\SCS(T, S_3 \mid S_1)$ & $\{a,o\}$ & $\{c,i,j,l,p,t\}$ & $\{b,c,d,h,i,m,p,w\}$ & $\{c,i,p\}$ & $0.433$ \\[2pt]
$\SCS(T, S_3 \mid S_2)$ & $\{a,p,t\}$ & $\{c,i,j,l,o\}$ & $\{b,c,d,h,i,m,w\}$ & $\{c,i\}$ & $0.338$ \\
\bottomrule
\end{tabular}
\caption{SCS scores for the synthetic $3 \to 1$ demonstration.}
\label{tab:scs-demo}
\end{table}

\subsection{Asymmetry}

All three role-swap pairs illustrate that asymmetry is contingent. It may or may not hold depending on the composition of the document sets:
\begin{align}
  \SCS(T, S_1 \mid S_2) &= 0.204 \neq 0.289 = \SCS(T, S_2 \mid S_1) \notag\\
  \SCS(T, S_1 \mid S_3) &= 0.375 \neq 0.433 = \SCS(T, S_3 \mid S_1) \notag\\
  \SCS(T, S_2 \mid S_3) &= 0.338 = 0.338 = \SCS(T, S_3 \mid S_2) \notag
\end{align}
The third pair happens to be symmetric because $|U_{S_2|S_3}| = |U_{S_3|S_2}| = 5$ and both residuals contribute exactly two matching lemmas to the same target residual $T \setminus (S_2 \cap S_3)$, so the numerators and denominators coincide. Such coincidences are not guaranteed: the asymmetry result (Section~\ref{sec:asymmetry}) states that $\SCS(T, R \mid C) \neq \SCS(T, C \mid R)$ \emph{in general}, not that it holds for every triple.

\subsection{Ranking reversal}

The average conditional scores and the resulting ranking are:
\begin{align}
  \overline{\SCS}(T, S_1) &= \tfrac{1}{2}(0.204 + 0.375) = 0.290 \notag\\
  \overline{\SCS}(T, S_2) &= \tfrac{1}{2}(0.289 + 0.338) = 0.313 \notag\\
  \overline{\SCS}(T, S_3) &= \tfrac{1}{2}(0.433 + 0.338) = 0.386 \notag
\end{align}

\begin{center}
\begin{tabular}{lcc}
\toprule
& STS ranking & SCS ranking \\
\midrule
1st & $S_2$ (0.375) & $S_3$ (0.386) \\
2nd & $S_3$ (0.375) & $S_2$ (0.313) \\
3rd & $S_1$ (0.335) & $S_1$ (0.290) \\
\bottomrule
\end{tabular}
\end{center}

The conditioning operation reverses the top of the ranking. The mechanism is transparent: $S_2$'s unconditioned similarity to $T$ was inflated by the lemmas $m$ and $w$, which $S_2$ shares with $S_1$. Once this lateral overlap is subtracted, $S_2$'s residual contribution to $T$ drops. $S_3$, by contrast, shares little with the other sources ($|S_3 \cap S_1| = 2$, $|S_3 \cap S_2| = 3$), so its overlap with $T$ is almost entirely its own. The conditioning operation confirms this by ranking $S_3$ first: it is the source whose content in the target is least attributable to vocabulary shared with other sources.

This is exactly the confound that motivates SCS: when sources share substantial lateral vocabulary, unconditioned STS cannot distinguish a source that contributes uniquely from one that merely shares the common pool. SCS resolves the confound by subtracting the shared pool before measuring alignment.

\subsection{Vertical role reversal: the $1 \to n$ case}

The same four sets can demonstrate the $1 \to n$ case by reassigning roles. Let $S_1$ serve as the single source~$S$, and let the remaining sets become three targets: $T_1 = S_2$, $T_2 = S_3$, $T_3 = T$. The analytical question reverses: not ``which source contributed most uniquely to the target?'' but ``which target uniquely received the most from the source?''

The unconditioned scores are:
\begin{align}
  \sigma(S, T_1) &= \frac{6}{\sqrt{10 \cdot 8}} = 0.671, \quad
  \sigma(S, T_2)  = \frac{2}{\sqrt{10 \cdot 8}} = 0.224, \quad
  \sigma(S, T_3)  = \frac{3}{\sqrt{10 \cdot 8}} = 0.335 \notag
\end{align}
giving the unconditioned ranking $T_1 > T_3 > T_2$.

The lateral overlaps are now among targets: $|T_1 \cap T_2| = 3$, $|T_1 \cap T_3| = 3$, $|T_2 \cap T_3| = 3$---moderately balanced, unlike the heavily skewed source-side overlaps in the $n \to 1$ case.

Table~\ref{tab:scs-1ton} reports the six $1 \to n$ SCS scores, computed via the role-reversed formula of Section~\ref{sec:one-to-n}.

\begin{table}[ht!]
\centering
\small
\begin{tabular}{@{}llllcc@{}}
\toprule
SCS$(S, R \mid C)$ & $I_{T_iT_j}$ & $U_{T_i|T_j}$ & $S \setminus I_{T_iT_j}$ & Overlap & Score \\
\midrule
$\SCS(S, T_1 \mid T_2)$ & $\{a,p,t\}$ & $\{f,m,q,w,y\}$ & 9 letters & $\{f,m,q,w,y\}$ & $0.745$ \\[2pt]
$\SCS(S, T_1 \mid T_3)$ & $\{m,p,w\}$ & $\{a,f,q,t,y\}$ & 8 letters & $\{a,f,q,y\}$ & $0.632$ \\[2pt]
$\SCS(S, T_2 \mid T_1)$ & $\{a,p,t\}$ & $\{c,i,j,l,o\}$ & 9 letters & $\{o\}$ & $0.149$ \\[2pt]
$\SCS(S, T_2 \mid T_3)$ & $\{c,i,p\}$ & $\{a,j,l,o,t\}$ & 10 letters & $\{a,o\}$ & $0.283$ \\[2pt]
$\SCS(S, T_3 \mid T_1)$ & $\{m,p,w\}$ & $\{b,c,d,h,i\}$ & 8 letters & $\{b\}$ & $0.158$ \\[2pt]
$\SCS(S, T_3 \mid T_2)$ & $\{c,i,p\}$ & $\{b,d,h,m,w\}$ & 10 letters & $\{b,m,w\}$ & $0.424$ \\
\bottomrule
\end{tabular}
\caption{SCS scores for the $1 \to n$ demonstration (vertical role reversal).}
\label{tab:scs-1ton}
\end{table}

The average scores are $\overline{\SCS}(S, T_1) = 0.689$, $\overline{\SCS}(S, T_2) = 0.216$, $\overline{\SCS}(S, T_3) = 0.291$, preserving the ranking $T_1 > T_3 > T_2$. In this instance the conditioning does not reverse the ranking, but the asymmetry is striking: $\SCS(S, T_1 \mid T_2) = 0.745$ versus $\SCS(S, T_2 \mid T_1) = 0.149$. Almost all of $T_1$'s residual content (after removing what it shares with~$T_2$) aligns with~$S$, while almost none of $T_2$'s residual content does. This is the $1 \to n$ analogue of the source-side confound: $T_1$ genuinely received the most from~$S$, and conditioning on other targets confirms rather than overturns that finding.

\subsection{Bilateral conditioning: the $m \to n$ case}

The bilateral case requires conditioning on \emph{both} sides of the source--target boundary simultaneously. Using $S_1$ and $S_2$ as the two sources and $S_3$ and $T$ as the two targets (relabeled $T_1$ and $T_2$), the source-side lateral overlap is $|S_1 \cap S_2| = 6$ and the target-side lateral overlap is $|T_1 \cap T_2| = 3$.

The source and target residuals are:
\begin{align}
  U_{S_1|S_2} &= \{b, e, k, o\} & (4~\text{letters}), &\qquad
  U_{S_2|S_1}  = \{p, t\} & (2~\text{letters}) \notag\\
  U_{T_1|T_2} &= \{a, j, l, o, t\} & (5~\text{letters}), &\qquad
  U_{T_2|T_1}  = \{b, d, h, m, w\} & (5~\text{letters}) \notag
\end{align}

\paragraph{Source-conditioned matrix $M_S$.}
Each entry conditions only on sources, using the $n \to 1$ formula with the source residuals $U_{S_1|S_2}$ and $U_{S_2|S_1}$. The target enters unconditioned but with the source lateral overlap $I_{S_1S_2} = \{a,f,m,q,w,y\}$ removed.

\begin{table}[ht!]
	\centering
	\small
	\begin{tabular}{@{}llllcc@{}}
		\toprule
		Entry & $U_{S_i|S_j}$ & $T_j \setminus I_{S_1S_2}$ & Overlap & Score \\
		\midrule
		$M_S[S_1, T_1]$ & $\{b,e,k,o\}$~(4) & $\{c,i,j,l,o,p,t\}$~(7) & $\{o\}$~(1) & $0.189$ \\[2pt]
		$M_S[S_1, T_2]$ & $\{b,e,k,o\}$~(4) & $\{b,c,d,h,i,p\}$~(6) & $\{b\}$~(1) & $0.204$ \\[2pt]
		$M_S[S_2, T_1]$ & $\{p,t\}$~(2) & $\{c,i,j,l,o,p,t\}$~(7) & $\{p,t\}$~(2) & $0.535$ \\[2pt]
		$M_S[S_2, T_2]$ & $\{p,t\}$~(2) & $\{b,c,d,h,i,p\}$~(6) & $\{p\}$~(1) & $0.289$ \\
		\bottomrule
	\end{tabular}
	\caption{Source-conditioned matrix $M_S$ for the $2 \times 2$ bilateral demonstration.}
	\label{tab:scs-ms}
\end{table}

\paragraph{Target-conditioned matrix $M_T$.}
Each entry conditions only on targets, using the $1 \to n$ formula with the target residuals $U_{T_1|T_2}$ and $U_{T_2|T_1}$. The source enters unconditioned but with the target lateral overlap $I_{T_1T_2} = \{c,i,p\}$ removed.

\begin{table}[ht!]
	\centering
	\small
	\begin{tabular}{@{}llllcc@{}}
		\toprule
		Entry & $U_{T_j|T_l}$ & $S_i \setminus I_{T_1T_2}$ & Overlap & Score \\
		\midrule
		$M_T[S_1, T_1]$ & $\{a,j,l,o,t\}$~(5) & $\{a,b,e,f,k,m,o,q,w,y\}$~(10) & $\{a,o\}$~(2) & $0.283$ \\[2pt]
		$M_T[S_1, T_2]$ & $\{b,d,h,m,w\}$~(5) & $\{a,b,e,f,k,m,o,q,w,y\}$~(10) & $\{b,m,w\}$~(3) & $0.424$ \\[2pt]
		$M_T[S_2, T_1]$ & $\{a,j,l,o,t\}$~(5) & $\{a,f,m,q,t,w,y\}$~(7) & $\{a,t\}$~(2) & $0.338$ \\[2pt]
		$M_T[S_2, T_2]$ & $\{b,d,h,m,w\}$~(5) & $\{a,f,m,q,t,w,y\}$~(7) & $\{m,w\}$~(2) & $0.338$ \\
		\bottomrule
	\end{tabular}
	\caption{Target-conditioned matrix $M_T$ for the $2 \times 2$ bilateral demonstration.}
	\label{tab:scs-mt}
\end{table}

\paragraph{Bilaterally conditioned matrix $M_B$.}
Each entry uses both the source and target residuals simultaneously.

\begin{table}[ht!]
	\centering
	\small
	\begin{tabular}{@{}lllcc@{}}
		\toprule
		Entry & $U_{S_i|S_j}$ & $U_{T_j|T_l}$ & Overlap & Score \\
		\midrule
		$M_B[S_1, T_1]$ & $\{b,e,k,o\}$~(4) & $\{a,j,l,o,t\}$~(5) & $\{o\}$~(1) & $0.224$ \\[2pt]
		$M_B[S_1, T_2]$ & $\{b,e,k,o\}$~(4) & $\{b,d,h,m,w\}$~(5) & $\{b\}$~(1) & $0.224$ \\[2pt]
		$M_B[S_2, T_1]$ & $\{p,t\}$~(2) & $\{a,j,l,o,t\}$~(5) & $\{t\}$~(1) & $0.316$ \\[2pt]
		$M_B[S_2, T_2]$ & $\{p,t\}$~(2) & $\{b,d,h,m,w\}$~(5) & $\emptyset$~(0) & $\mathbf{0.000}$ \\
		\bottomrule
	\end{tabular}
	\caption{Bilaterally conditioned matrix $M_B$ for the $2 \times 2$ bilateral demonstration.}
	\label{tab:scs-mb}
\end{table}

\paragraph{Summary comparison.} Table~\ref{tab:scs-mton} compares the unconditioned STS matrix with the three conditioned matrices defined in Section~\ref{sec:m-to-n}: source-conditioned~$M_S$, target-conditioned~$M_T$, and bilaterally conditioned~$M_B$.

\begin{table}[ht!]
	\centering
	\small
	\begin{tabular}{@{}lcccc@{}}
		\toprule
		Pair & STS & $M_S$ & $M_T$ & $M_B$ \\
		\midrule
		$(S_1, T_1)$ & $0.224$ & $0.189$ & $0.283$ & $0.224$ \\
		$(S_1, T_2)$ & $0.335$ & $0.204$ & $0.424$ & $0.224$ \\
		$(S_2, T_1)$ & $0.375$ & $0.535$ & $0.338$ & $0.316$ \\
		$(S_2, T_2)$ & $0.375$ & $0.289$ & $0.338$ & $\mathbf{0.000}$ \\
		\bottomrule
	\end{tabular}
	\caption{Four-matrix comparison for the $2 \times 2$ bilateral demonstration. $M_S$: source-conditioned; $M_T$: target-conditioned; $M_B$: bilaterally conditioned.}
	\label{tab:scs-mton}
\end{table}

The most revealing entry is $M_B[S_2, T_2] = 0$. Sources $S_2$ and $T_2$ have an unconditioned STS of~$0.375$, a source-conditioned score of~$0.289$, and a target-conditioned score of~$0.338$---yet their \emph{bilateral} score is zero. The reason is visible in the residuals: $U_{S_2|S_1} = \{p, t\}$ and $U_{T_2|T_1} = \{b, d, h, m, w\}$, which share no letters at all. Everything $S_2$ contributes to~$T_2$ is content that $S_2$ shares with $S_1$ or that $T_2$ shares with $T_1$ (or both). Bilateral conditioning strips away both lateral pools, revealing that the $S_2$--$T_2$ link is entirely parasitic on shared vocabulary.

A noteworthy property of the bilateral case is that $M_B$ is \emph{not} bounded above by $\min(M_S, M_T)$. For instance, $M_B[S_1, T_1] = 0.224 > M_S[S_1, T_1] = 0.189$. This occurs because cosine normalization divides by~$\sqrt{|U_S| \cdot |U_T|}$: when bilateral conditioning reduces the target residual from~8 to~5 letters without reducing the 1-letter overlap, the smaller denominator ($\sqrt{20}$ versus $\sqrt{28}$) pushes the score \emph{up}. The residual \emph{sets} are subsets of the originals; the cosine \emph{scores} need not be.

%------------------------------------------------------------------
\section{Related Work and Positioning}
\label{sec:related}

\subsection{Tversky's contrast model and the asymmetry of similarity}
\label{sec:tversky}

\citet{tversky1977features} challenged the then-dominant geometric models of similarity by arguing that similarity judgments are better explained as a feature-matching process than as a distance in metric space. His \emph{contrast model} defines similarity as a linear combination of the measures of common and distinctive features:
\[
  S(a,b) \;=\; \theta\, f(A \cap B) \;-\; \alpha\, f(A \setminus B) \;-\; \beta\, f(B \setminus A),
\]
where $A$ and $B$ are the feature sets of the two stimuli, $f$ is a monotone set function, and $\theta$, $\alpha$, $\beta$ are non-negative weights. The central empirical finding is that subjects weight distinctive features asymmetrically: a prominent stimulus used as the referent contributes more weight to the ``$A \setminus B$'' term than when the roles are reversed, so that $S(a,b) \neq S(b,a)$---a direct violation of the symmetry axiom that metric-space similarity theories presuppose.

SCS shares two of Tversky's premises and departs on a third. Both frameworks (i)~decompose similarity into shared and distinctive components and (ii)~predict asymmetry as a structural consequence of that decomposition. The departure is in the \emph{source} of asymmetry. In the contrast model, asymmetry arises from differential \emph{weighting} of two-way distinctive features---the same sets receive different coefficients depending on which stimulus is the referent. In SCS, asymmetry arises from differential \emph{content}: the conditioning operation produces different residual sets when the roles of reference and conditioning document are exchanged (Section~\ref{sec:asymmetry}), and the similarity function applied to those residuals is itself symmetric.

The contrast model operates on pairs and introduces asymmetry through weighting; SCS operates on triples and introduces asymmetry through conditioning. Tversky's insight that similarity need not be symmetric is the intellectual ancestor; SCS provides a mechanism by which asymmetry emerges without free parameters, from the set-theoretic structure of a multi-document configuration.

\subsection{Conditional similarity: from vision to language}
\label{sec:csts-positioning}

The insight that a single unconditional notion of similarity is insufficient when multiple aspects of resemblance coexist was first formalized in computer vision. \citet{veit2017conditional} introduced Conditional Similarity Networks (CSNs), which learn disentangled embedding subspaces, each capturing a different aspect of image similarity (e.g., color, shape, category); a learned mask selects the relevant subspace for a given condition. \citet{shi2025generalized} extended this framework to the weakly supervised setting, where the condition label for each comparison is unknown and must be inferred via semantic matching. In these vision approaches, the condition selects \emph{which similarity function} to apply; in SCS, a single similarity function operates on inputs whose lexical content has been modified by set-theoretic operations with the conditioning document.

In NLP, \citet{deshpande2023csts} introduced Conditional Semantic Textual Similarity (\CSTS{}) at EMNLP 2023, proposing a task in which the similarity between two sentences is evaluated with respect to a natural-language condition describing an aspect of interest. Subsequent work extended \CSTS{} along architectural and training dimensions: \citet{zhang2026case} proposed Condition-Aware Sentence Embeddings (CASE); \citet{song2025polirl} proposed PoLi-RL; \citet{liu2025conditional} proposed a conditional contrastive learning framework; and earlier contributions include \citet{li2024seaver}, \citet{zhang2025annotating}, \citet{li2024lcsts}, and \citet{yamada2025unsupervised}. Despite the shared use of the term ``conditional,'' SCS and the \CSTS{} family are fundamentally different operations. The distinction is best understood through three progressively deeper structural viewpoints.

\paragraph{Distinction 1: Scope.} The entire C-STS line of research operates at the sentence level with pairwise comparison. SCS operates at the document level with three or more documents simultaneously in an $n \to 1$ configuration. The shift from pairwise comparison to multi-document decomposition changes the structure of the problem from evaluating a single relationship to decomposing a system of relationships.

\paragraph{Distinction 2: Conditioning mechanism.} In C-STS and all its descendants, the condition is a compound noun phrase that names a semantic attribute of the entities described in the sentences (e.g., ``the color of the object''). This is \emph{intra-document conditioning}: the condition directs attention to aspects internal to the documents being compared. In SCS, the condition is another document whose lexical intersection with the document under analysis is operationally subtracted. This is \emph{inter-document conditioning}: the condition is a concrete textual entity whose shared content is removed to isolate what remains. The CASE subtraction operation---subtracting the unconditional condition embedding---is a noteworthy parallel: both CASE and SCS employ an ``eliminate-then-compare'' strategy, but they eliminate different things for different reasons.

\paragraph{Distinction 3: Partitioning.} C-STS produces similarity scores on a continuous Likert scale (1--5) with no principled boundary between similar and dissimilar. The standard evaluation metric---Spearman rank correlation---sidesteps the partitioning question. SCS operates through set operations that produce a lexical decomposition requiring no threshold. A lemma is either in the intersection or it is not. This is a direct consequence of Contrast~2: precisely because SCS conditions via set operations, it can partition without thresholds.

\paragraph{Mathematical consequences.} The cumulative structural differences produce two mathematical properties absent from C-STS: \emph{directionality} (Section~\ref{sec:directionality}) and \emph{asymmetry} (Section~\ref{sec:asymmetry}). C-STS preserves symmetry.

\paragraph{Complementarity.} The two frameworks address different 
questions: C-STS asks ``similar \emph{in what respect}?''; SCS asks 
``similar \emph{after removing what}?'' Whether they can be composed  in a single pipeline---applying SCS-style subtraction first to isolate marginal content, then C-STS-style aspect selection to evaluate the residual---remains an open question, since the two operate at different granularities (sentence vs.\ document) and use fundamentally different conditioning mechanisms (intra- vs.\ inter-document).

\subsection{Positioning with respect to the duality of similarity and metric spaces}
\label{sec:rozinek-positioning}

\citet{rozinek2021duality} established a rigorous set-theoretic foundation for the relationship between similarity and distance, deriving generally valid conversion relations and showing that measures such as Jaccard, Tanimoto, Steinhaus, and others satisfy a unified axiom system.

SCS accepts and builds upon this set-theoretic foundation. However, SCS reveals that the metric space axioms---which \citeauthor{rozinek2021duality} inherit and preserve throughout---function not merely as structural guarantees but also as constraints that limit the expressiveness of the formalism. The symmetry axiom encodes a substantive assumption that the similarity between two documents is independent of which document is taken as the reference. In their unified theory, this assumption is never relaxed, and consequently the framework characterizes only the \emph{unconditional, symmetric} case.

SCS demonstrates, by construction, that this assumption becomes untenable in multi-document settings where documents occupy role-differentiated positions. The conditioning operation---fully compatible with the foundations of \citeauthor{rozinek2021duality}---produces role-induced asymmetry (Section~\ref{sec:asymmetry}) without relaxing the similarity function itself.

Symmetry is an appropriate axiom for unordered-pair similarity. It limits what the formalism can express when documents occupy distinct roles. SCS extends the domain to ordered, conditioned triples in which role assignment changes the residualized inputs, identifying a class of conditional relationships that symmetric axioms do not reach. This does not invalidate symmetric similarity spaces; \citeauthor{rozinek2021duality} unified one side of the story; SCS opens the other.

\subsection{Comparative document analysis}
\label{sec:comparative-doc}

The decomposition of document content into shared and unique components has precedent in comparative document analysis, where it serves summarization rather than similarity measurement. \citet{mani1999summarizing} proposed a graph-based approach using spreading activation to separate common from unique concepts across related documents, enabling summaries that highlight both similarities and differences. \citet{ren2017comparative} extended this direction at scale, decomposing documents into shared and unique phrase sets for contrastive summarization of large text corpora. Both lines of work recognize the same structural insight that underlies SCS---that the relationship between documents is richer than a single scalar score can express, and that separating shared from unique content is analytically productive. However, they apply the decomposition to \emph{generate text} (summaries of similarities and differences), whereas SCS applies it to \emph{measure similarity} (a conditional score that isolates marginal contribution). The shared/unique decomposition is the common substrate; the downstream task distinguishes the two research programs.

%------------------------------------------------------------------
\section{Discussion}
\label{sec:discussion}

\subsection{Symmetry as a constraint on understanding}

As Section~\ref{sec:asymmetry} demonstrated, the role-induced asymmetry of SCS arises from the inputs, not the function. The pairwise symmetry axiom is therefore not a flaw in classical STS but a faithful reflection of the pairwise setting; it becomes a constraint only when the comparison domain extends to role-differentiated configurations.

A related open question is whether SCS's inter-document conditioning 
and C-STS's intra-document conditioning (Section~\ref{sec:csts-positioning}) can be composed, given their differences in granularity and mechanism.

\subsection{Term frequency and the conditioning mechanism}
\label{sec:tf-discussion}

In traditional unconditioned pairwise STS, term frequency exerts a systematic influence on similarity scores. High-frequency domain terms---words that recur across most documents in a given field---contribute disproportionately to every pairwise comparison, inflating all scores and obscuring the differential alignment that is often the object of inquiry. In the motivating application of this paper, foundational ethical vocabulary such as ``harm,'' ``fairness,'' ``transparency,'' and ``accountability'' pervades all source theories; their shared presence elevates every theory's similarity to any target text, regardless of whether the alignment reflects a genuine unique contribution.

Term Frequency–Inverse Document Frequency (TF-IDF) weighting is the 
standard remedy: it downweights terms that are frequent across a corpus and amplifies rare, distinctive ones. However, TF-IDF operates on corpus-wide distributional statistics---inverse document frequency is a property of the term's prevalence across \emph{all} documents in the collection, not of its role in the specific comparison at hand. A term that is rare corpus-wide but happens to be shared by both the reference and the conditioning document still receives a high IDF weight, even though it carries no differentiating information for the triple under analysis.

SCS addresses the same confound through a fundamentally different mechanism. Rather than \emph{reweighting} shared terms, it \emph{removes} them: the conditioning operation subtracts the lateral shared context $I_{RC} = R \cap C$ from both the reference and the target before similarity is computed. The high-frequency domain terms that inflate traditional STS scores are precisely the ones most likely to appear in multiple source documents and therefore to fall into the intersection that conditioning eliminates. What remains in the residual sets $U_{R|C}$ and $T \setminus (R \cap C)$ are the lower-frequency, distinctive terms that carry the actual differentiating signal---the terms that TF-IDF weighting would amplify but cannot isolate from the shared context.

The two approaches are complementary rather than competing. Frequency weighting adjusts the \emph{magnitude} of each term's contribution; SCS conditioning adjusts the \emph{membership} of the term sets entering the comparison. A hybrid approach---applying TF-IDF weighting to the residual sets after conditioning---would combine set-theoretic removal of shared context with distributional reweighting of the remaining terms. However, such an extension would replace the binary indicator vectors with real-valued vectors, and the clean set-cardinality expressions of  Definition~\ref{def:cosine-scs} and Equation~\ref{eq:cosine-scs-set} would no longer hold. The trade-off between the interpretive transparency of binary representation and the potential discriminative gain of frequency weighting is an empirical question for future work.

\subsection{Scope and limitations}

Several limitations should be acknowledged. Binary vectorization discards frequency information, a design choice whose rationale and trade-offs are discussed in Section~\ref{sec:tf-discussion}; extension to TF-IDF weighting is algebraically straightforward but may affect the asymmetry result. The quality of the decomposition depends on the quality of the lemmatization. An alternative using dense embeddings rather than sparse binary vectors would capture semantic similarity beyond lexical co-occurrence but would sacrifice threshold-free partitioning.

The set-theoretic operations that underlie SCS place implicit
requirements on the documents to which it is applied.
“Because conditioning subtracts the intersection of two lemma sets before computing cosine similarity on the residual, documents must be long enough to produce dense, stable term-frequency estimates; very short texts (a single sentence or a tweet) yield sparse vectors whose residuals are dominated by noise.  

In practice, documents of at least several hundred
words—and ideally one thousand or more—provide a reliable
substrate for the intersection and subtraction steps.
Equally important is the degree of thematic overlap among
documents in the corpus: if two documents share almost all of
their vocabulary, conditioning removes nearly everything and
the residual collapses toward zero; if they share almost
nothing, conditioning removes nothing and SCS reduces to
ordinary STS.  The method is therefore most informative in the
middle ground—corpora whose documents share partial thematic
overlap yet retain substantial unique content.

Domains that naturally exhibit this structure include
regulatory and legal texts addressing related topics,
academic papers within a subfield, news coverage of the same
event from multiple outlets, and corporate reports from the
same industry.  The synthetic letter-set demonstrations in
Section~\ref{sec:demonstration} were designed to make the
set-theoretic mechanics transparent; validating the framework
on natural-language corpora with the properties described above
is a natural direction for future work.

\subsection{Future directions}

SCS was motivated by ethical influence analysis but is domain-agnostic. 
The four cardinality configurations suggest natural applications: 
$n \to 1$ for legal precedent analysis and patent prior-art retrieval, 
where multiple source documents are compared against a single target; 
$1 \to n$ for tracing a single source's differential reach across 
multiple targets, as in statutory influence; and $m \to n$ for 
settings where both sides are plural, such as policy diffusion 
across jurisdictions or literary intertextuality. Testing 
generalizability across these domains and configurations is a 
primary empirical direction.

At large $k$ or $m \times n$, clustering of the SCS score matrices 
may identify groups of documents with similar contribution profiles; 
the threshold for viability is an open empirical question. A hybrid 
approach using set-theoretic conditioning to define subsets, then 
computing similarity via transformer-based embeddings of those 
subsets, could combine the set-theoretic advantages of SCS with the 
semantic depth of contextual embeddings.

%------------------------------------------------------------------
\section{Conclusion}
\label{sec:conclusion}

This paper introduced Semantic Conditional Similarity (SCS), a conditional form of STS in which the similarity between two documents is computed after eliminating semantic elements shared with a third document. SCS makes four contributions: it introduces the concept of \emph{directionality} to semantic similarity; it demonstrates that extending the comparison domain to ordered, conditioned triples induces \emph{role-dependent asymmetry} as a mathematical consequence; it introduces \emph{inter-document conditioning} as a complement to the intra-document conditioning of C-STS; and it achieves \emph{threshold-free partitioning} through set-theoretic operations. Although developed for the $n \to 1$ configuration, the framework extends to the $1 \to n$ case through a vertical role reversal and to the $m \to n$ case through the independent composition of source-side and target-side conditioning. Because the two conditioning operations act on disjoint document sets, they compose without interaction, and all four contributions carry over to the bilateral case. Together with the established $1 \to 1$ case, SCS thus provides a complete family of conditional similarity measures covering the full document cardinality landscape.

SCS scores should be interpreted ordinally---as providing an ordering of marginal contributions---rather than as absolute measurements. Natural language is inherently ambiguous and vague; any procedure that maps it to a real-valued similarity score performs a substantial reduction, compressing contextual, pragmatic, and connotative dimensions into a single number. Because SCS rankings depend only on the \emph{order} of the scores, not on their exact magnitudes, they are robust to monotone distortions of the scale---making the ranking valid even when the binary indicator vectors provide only an approximate numerical representation. Rankings ask only that the mapping preserve order, not magnitude, and are thus the epistemologically appropriate level of interpretation for a framework built on set-cardinality proxies for semantic content.

The broader implication is that the metric space axioms commonly assumed in similarity formalizations are not neutral structural scaffolding but substantive claims about the nature of the relationships being modeled. When the application domain involves role differentiation among documents, the pairwise symmetric framework suppresses information that an extended, conditioned comparison domain can recover. SCS makes this visible, and in doing so, opens a side of the similarity--distance duality that the symmetric, unconditional case leaves closed.

%------------------------------------------------------------------
\section*{Acknowledgments}

The author thanks Prof.\ Mehmet Nafiz Aydin for supervision of the doctoral research from which this work derives. The author is particularly grateful to Assoc.\ Prof.\ Emrullah Fatih Yetkin for emphasizing the importance of theoretical grounding and validation during the progression of the doctoral research. The author also thanks the members of the doctoral jury for their examination and feedback during the defense.

%------------------------------------------------------------------
\section*{Ethics Statement}

This work is purely theoretical and mathematical: it introduces formal definitions, theorems, and proofs within the framework of set-theoretic similarity analysis. Neither human subjects, nor information about individuals, groups, or organizations are used. 

%------------------------------------------------------------------
\section*{Data Availability Statement}

This paper presents a theoretical mathematical framework and contains neither empirical nor external datasets. The synthetic demonstration in Section~\ref{sec:demonstration} uses randomly generated letter sets (Python, seed 5663). 

%------------------------------------------------------------------
\bibliographystyle{acl_natbib}
\bibliography{references}

\end{document}